\documentclass[11pt, a4paper]{lumia}
\pdfoutput=1
\usepackage[sort&compress]{natbib}
\usepackage{xspace}

\theoremstyle{plain}

\newtheorem*{proposition*}{Proposition}

\theoremstyle{definition}

\theoremstyle{definition}

\def\eqref#1{equation~\ref{#1}}

\usepackage{graphicx}           
\usepackage{tikz}               
\usepackage[edges]{forest}      

\usepackage{url}                
\usepackage{xurl}               

\usepackage{array}              
\usepackage{longtable}          
\usepackage{multirow}           
\usepackage{makecell}           
\usepackage{ragged2e}           

\usepackage{mathtools}          
\usepackage{nicefrac}           

\usepackage{algorithm}          
\usepackage{algorithmicx}       
\usepackage{algpseudocode}      
\usepackage{listings}           

\usepackage{subcaption}         
\usepackage{wrapfig}            
\usepackage[export]{adjustbox}  

\usepackage{changes}            
\usepackage{xspace}             
\usepackage[normalem]{ulem}     
\usepackage{CJKutf8}            

\usepackage[tikz]{bclogo}       
\usepackage[framemethod=tikz]{mdframed} 

\usepackage{placeins}
\usepackage{cleveref}
\usepackage{lipsum}             
\usepackage{tocloft}            
\usepackage{afterpage}          
\usepackage{bbding}             
\usepackage{epigraph}           
\usepackage{minitoc}            
\usepackage{multicol}           
\usepackage{textgreek}          
\newcommand{\ourmethod}{{\fontfamily{lmtt}\selectfont \textbf{LatentMem}}\xspace}
\newcommand{\llmname}[1]{{\fontfamily{pcr}\selectfont {#1}}\xspace}

\newcommand{\hlfirst}[1]{%
  \begingroup
  \setlength{\fboxsep}{1pt}%
  \colorbox[HTML]{CFE2FF}{#1}%
  \endgroup
}

\newcommand{\hlsecond}[1]{%
  \begingroup
  \setlength{\fboxsep}{1pt}%
  \colorbox[HTML]{E8F1FF}{#1}%
  \endgroup
}

\newcommand{\obsbox}[1]{%
    \begin{tcolorbox}[colframe=black!70, colback=yellow!5, boxrule=1pt, arc=1mm,   top=3pt, bottom=3pt, left=3pt, right=3pt,
  boxsep=1pt]
        #1
    \end{tcolorbox}
}

\definecolor{mygreen}{rgb}{0.29, 0.7, 0.48}
\definecolor{darksalmon}{rgb}{0.91, 0.59, 0.48}
\definecolor{mygrey}{gray}{0.4}
\newcommand{\degrade}[1]{$_{\color{RedOrange}\downarrow #1}$}
\newcommand{\upgrade}[1]{$_{\color{JungleGreen}\uparrow #1}$}
\usepackage[most]{tcolorbox}
\usepackage{makecell}
\newcolumntype{P}[1]{>{\RaggedRight\arraybackslash}p{#1}}

\definecolor{uclablue}{RGB}{39, 116, 174}
\definecolor{bigaired}{RGB}{156, 0, 0}
\definecolor{myblue}{HTML}{598BE7}
\definecolor{mildblue}{RGB}{31,119,180}
\definecolor{sectionblue}{RGB}{70, 130, 180}
\definecolor{methodblue}{RGB}{0, 150, 136}
\definecolor{bgblue}{RGB}{245,243,253}
\definecolor{ttblue}{RGB}{91,194,224}
\definecolor{mygreen}{rgb}{0.64, 0.56, 0.88}
\definecolor{myyellow}{rgb}{0.68, 0.6, 0.1}
\definecolor{fancygreen}{rgb}{0.33, 0.68, 0.20}
\definecolor{salmon}{rgb}{0.94, 0.52, 0.49}
\definecolor{tablegreen}{rgb}{0.82, 0.94, 0.75}
\definecolor{tableblue}{rgb}{0.81, 0.90, 0.94}
\definecolor{tablered}{rgb}{0.97, 0.85, 0.85}
\definecolor{tableorange}{rgb}{0.96, 0.85, 0.81}
\definecolor{myorange}{rgb}{1.0, 0.49, 0.0}
\definecolor{tlgreen}{rgb}{0.33, 0.68, 0.20}
\definecolor{darkgreen}{RGB}{0,100,0}
\definecolor{darkred}{RGB}{200, 0, 0}
\definecolor{customyellow}{HTML}{FFFACD}
\definecolor{refinegreen}{RGB}{0, 128, 75}
\definecolor{scoregreen}{RGB}{34, 139, 34}
\definecolor{hidden-blue}{RGB}{194,232,247}
\definecolor{hidden-black}{RGB}{20,68,106}
\definecolor{yes}{HTML}{C6EFCE}
\definecolor{no}{HTML}{FFC7CE}
\definecolor{partial}{HTML}{FFEB9C}
\definecolor{external}{HTML}{D9E1F2}
\definecolor{hdr}{HTML}{F2F2F2}
\definecolor{GRPOrow}{gray}{0.96}
\definecolor{FlowRLrow}{RGB}{225,236,255}
\definecolor{FlowBlue}{RGB}{80,120,210}
\definecolor{GRPOGray}{gray}{0.35}

\setlist[itemize]{leftmargin=20pt, noitemsep, topsep=0pt}

\NewDocumentCommand{\kaiyan}{mO{}}{\textcolor{purple}{\textsuperscript{\textit{kaiyan}}\textsf{\textbf{\small[#1]}}}}
\NewDocumentCommand{\yuxin}{mO{}}{\textcolor{cyan}{\textsuperscript{\textit{yuxin}}\textsf{\textbf{\small[#1]}}}}
\NewDocumentCommand{\bx}{mO{}}{\textcolor{green}{\textsuperscript{\textit{bx}}\textsf{\textbf{\small[#1]}}}}
\NewDocumentCommand{\at}{mO{}}{\textcolor{red}{\textsuperscript{\textit{AT}}\textsf{\textbf{\small[#1]}}}}
\NewDocumentCommand{\re}{mO{}}{\textcolor{blue}{\textsuperscript{\textit{RE}}\textsf{\textbf{\small[#1]}}}}
\NewDocumentCommand{\ybsun}{mO{}}{\textcolor{magenta}{\textsuperscript{\textit{youbang}}\textsf{\textbf{\small[#1]}}}}
\NewDocumentCommand{\runze}{mO{}}{\textcolor{orange}{\textsuperscript{\textit{runze}}\textsf{\textbf{\small[#1]}}}}
\NewDocumentCommand{\add}{mO{}}{\textcolor{darkgreen}{\textsuperscript{\textit{Maybe Consider Discuss}}\textsf{\textbf{[#1]}}}}

\newcommand{\cmark}{\textcolor{darkgreen}{\boldmath$\checkmark$}}
\newcommand{\xmark}{\textcolor{darkred}{\boldmath$\times$}}

\newenvironment{itemize*}%
 {\leftmargini=10pt\begin{itemize}%
  \setlength{\itemsep}{0pt}%
  \setlength{\parskip}{0pt}%
  }%
 {\end{itemize}}

\newenvironment{enumerate*}%
 {\begin{enumerate}%
  \setlength{\itemsep}{0pt}%
  \setlength{\parskip}{0pt}}%
 {\end{enumerate}}

\newcommand{\cellstatus}[1]{%
  \begingroup
  \StrTrim{#1}[\statusval]%
  \IfStrEq{\statusval}{Yes}{\cellcolor{yes}\cmark}{}%
  \IfStrEq{\statusval}{No}{\cellcolor{no}\xmark}{}%
  \IfBeginWith{\statusval}{Yes (}{\cellcolor{yes}\cmark~\textit{\statusval\unskip}}{}%
  \IfStrEq{\statusval}{Partial}{\cellcolor{partial}\textbf{Partial}}{}%
  \IfStrEq{\statusval}{External}{\cellcolor{external}\textbf{External}}{}%
  \endgroup
}

\newtcolorbox{myboxi}[1][]{
  breakable,
  title=#1,
  colback=red!5,
  colbacktitle=red!5,
  coltitle=black,
  fonttitle=\bfseries,
  bottomrule=0pt,
  toprule=0pt,
  leftrule=2pt,
  rightrule=2pt,
  titlerule=0pt,
  arc=0pt,
  outer arc=0pt,
  colframe=red,
}

\newtcolorbox{myboxnote}[1][]{
  breakable,
  title=#1,
  colback=orange!0,
  colbacktitle=orange!0,
  coltitle=black,
  fonttitle=\bfseries,
  bottomrule=0pt,
  toprule=0pt,
  leftrule=2pt,
  rightrule=2pt,
  titlerule=0pt,
  arc=0pt,
  outer arc=0pt,
  colframe=orange,
}

\newtcolorbox{myboxii}[1][]{
  breakable,
  freelance,
  title=#1,
  colback=white,
  colbacktitle=white,
  coltitle=black,
  fonttitle=\bfseries,
  bottomrule=0pt,
  boxrule=0pt,
  colframe=white,
  overlay unbroken and first={
  \draw[red!75!black,line width=3pt]
    ([xshift=5pt]frame.north west) -- 
    (frame.north west) -- 
    (frame.south west);
  \draw[red!75!black,line width=3pt]
    ([xshift=-5pt]frame.north east) -- 
    (frame.north east) -- 
    (frame.south east);
  },
  overlay unbroken app={
  \draw[red!75!black,line width=3pt,line cap=rect]
    (frame.south west) -- 
    ([xshift=5pt]frame.south west);
  \draw[red!75!black,line width=3pt,line cap=rect]
    (frame.south east) -- 
    ([xshift=-5pt]frame.south east);
  },
  overlay middle and last={
  \draw[red!75!black,line width=3pt]
    (frame.north west) -- 
    (frame.south west);
  \draw[red!75!black,line width=3pt]
    (frame.north east) -- 
    (frame.south east);
  },
  overlay last app={
  \draw[red!75!black,line width=3pt,line cap=rect]
    (frame.south west) --
    ([xshift=5pt]frame.south west);
  \draw[red!75!black,line width=3pt,line cap=rect]
    (frame.south east) --
    ([xshift=-5pt]frame.south east);
  },
}

\tcbset{
  takeawaysbox/.style={
    title=Takeaways,
    colback=lightblue!80,
    colframe=black,
    fonttitle=\bfseries\small,
    coltitle=white,
    colbacktitle=black,
    enhanced,
    attach boxed title to top left={xshift=2.5mm,yshift=-2.5mm},
    boxed title style={rounded corners, size=small, colframe=black, colback=black},
    width=\linewidth,
    arc=3.5mm
  }
}

\mdfdefinestyle{mystyle}{%
  rightline=true,
  innerleftmargin=10,
  innerrightmargin=10,
  outerlinewidth=3pt,
  topline=false,
  rightline=true,
  bottomline=false,
  skipabove=\topsep,
  skipbelow=\topsep
}

\tikzset{%
    every node/.style={font=\tiny},
    parent/.style =          {align=center,text width=2cm,rounded corners=3pt, line width=0.3mm, fill=gray!10,draw=gray!80},
    child/.style =           {align=center,text width=2.0cm,rounded corners=3pt, fill=blue!10,draw=blue!80,line width=0.3mm},
    grandchild/.style =      {align=center,text width=2cm,rounded corners=3pt},
    greatgrandchild/.style = {align=center,text width=1.5cm,rounded corners=3pt},
    greatgrandchild2/.style = {align=center,text width=1.5cm,rounded corners=3pt},    
    referenceblock/.style =  {align=center,text width=1.5cm,rounded corners=2pt},
    pretrain/.style =           {align=center,text width=2.0cm,rounded corners=3pt, fill=blue!10,draw=blue!80,line width=0.3mm},   
    pretrain_work/.style =           {align=center, text width=8.5cm,rounded corners=3pt, fill=blue!10,draw=blue!0,line width=0.3mm},  
    template/.style =           {align=center,text width=2.0cm,rounded corners=3pt, fill=red!10,draw=red!80,line width=0.3mm},   
    template_work/.style =           {align=center,text width=8.5cm,rounded corners=3pt, fill=red!10,draw=red!0,line width=0.3mm},    
    answer/.style =           {align=center,text width=2.0cm,rounded corners=3pt, fill= cyan!10,draw= cyan!80,line width=0.3mm},   
    answer_work/.style =           {align=center,text width=8.5cm,rounded corners=3pt, fill= cyan!10,draw= cyan!0,line width=0.3mm},      
    multiple/.style =           {align=center,text width=2.0cm,rounded corners=3pt, fill= orange!10,draw= orange!80,line width=0.3mm},   
    multiple_work/.style =           {align=center,text width=8.5cm,rounded corners=3pt, fill= orange!10,draw= orange!0,line width=0.3mm},        
    tuning/.style =           {align=center,text width=2.0cm,rounded corners=3pt, fill= magenta!10,draw= magenta!80,line width=0.3mm},   
    tuning_work/.style =           {align=center,text width=8.5cm,rounded corners=3pt, fill= magenta!10,draw= magenta!0,line width=0.3mm},          
}

\newcommand{\lstbg}[3][0pt]{{\fboxsep#1\colorbox{#2}{\strut #3}}}

\lstdefinelanguage{diff}{
  basicstyle=\ttfamily\small,
  morecomment=[f][\lstbg{red!20}]-,
  morecomment=[f][\lstbg{green!20}]+,
}

\lstdefinelanguage{diffpython}{
  language=diff,
  morekeywords={def, if, else, for, while, return, import, from, as, class, with, try, except, finally, raise, lambda, and, or, not, in, is, None, True, False},
  morecomment=[l]{\#},
  morestring=[b]",
  morestring=[b]',
}

\setheadertext{
    \includegraphics[height=1.2cm]{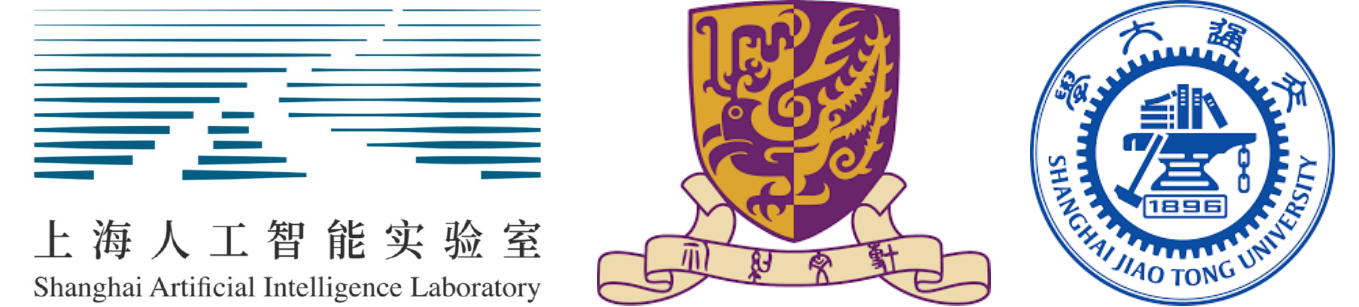}
}

\correspondingemail{\emailicon\ boonkana10@gmail.com
 \quad $^\ddagger$ Corresponding Author.}

\githublink {https://github.com/KANABOON1/LatentMem}

\huggingfacelink{https://huggingface.co/Kana-s/LatentMem-Qwen3-4B}

\renewcommand{\today}{2026-2-1}

\title{LatentMem: Customizing Latent Memory for Multi-Agent Systems}
\setheadertitle{LatentMem: Customizing Latent Memory for Multi-Agent Systems}

\author{
  Muxin Fu$^{1,2}$, Xiangyuan Xue$^{3}$, Yafu Li$^{3}$, Zefeng He$^{4}$, Siyuan Huang$^{5}$,  Xiaoye Qu$^{2\ddagger}$, Yu Cheng$^{3\ddagger}$, Yang Yang$^{5\ddagger}$\\
  $^1$Tongji University \quad $^2$Shanghai AI Laboratory \quad 
  $^3$The Chinese University of Hong Kong \quad\\
  $^4$Nanjing University \quad
  $^5$Shanghai Jiao Tong University
}

\begin{document}

\begin{abstract}
Large language model (LLM)-powered multi-agent systems (MAS) demonstrate remarkable collective intelligence, wherein multi-agent memory serves as a pivotal mechanism for continual adaptation.
However, existing multi-agent memory designs remain constrained by two fundamental bottlenecks: \emph{(i) memory homogenization} arising from the absence of role-aware customization, and \emph{(ii) information overload} induced by excessively fine-grained memory entries. To address these limitations, we propose \ourmethod, a learnable multi-agent memory framework designed to customize agent-specific memories in a token-efficient manner. 
Specifically, \ourmethod comprises an \textit{experience bank} that stores raw interaction trajectories in a lightweight form, and a \textit{memory composer} that synthesizes compact latent memories conditioned on retrieved experience and agent-specific contexts. Further, we introduce \textit{Latent Memory Policy Optimization} (LMPO), which propagates task-level optimization signals through latent memories to the composer, encouraging it to produce compact and high-utility representations.
Extensive experiments across diverse benchmarks and mainstream MAS frameworks show that \ourmethod achieves a performance gain of up to $19.36\%$ over vanilla settings and consistently outperforms existing memory architectures, without requiring any modifications to the underlying frameworks.
\end{abstract}

\maketitle

\section{Introduction}
\begin{wrapfigure}{r}{0.5\textwidth}
  \vspace{-1.0em}
  \centering
  \includegraphics[width=\linewidth]{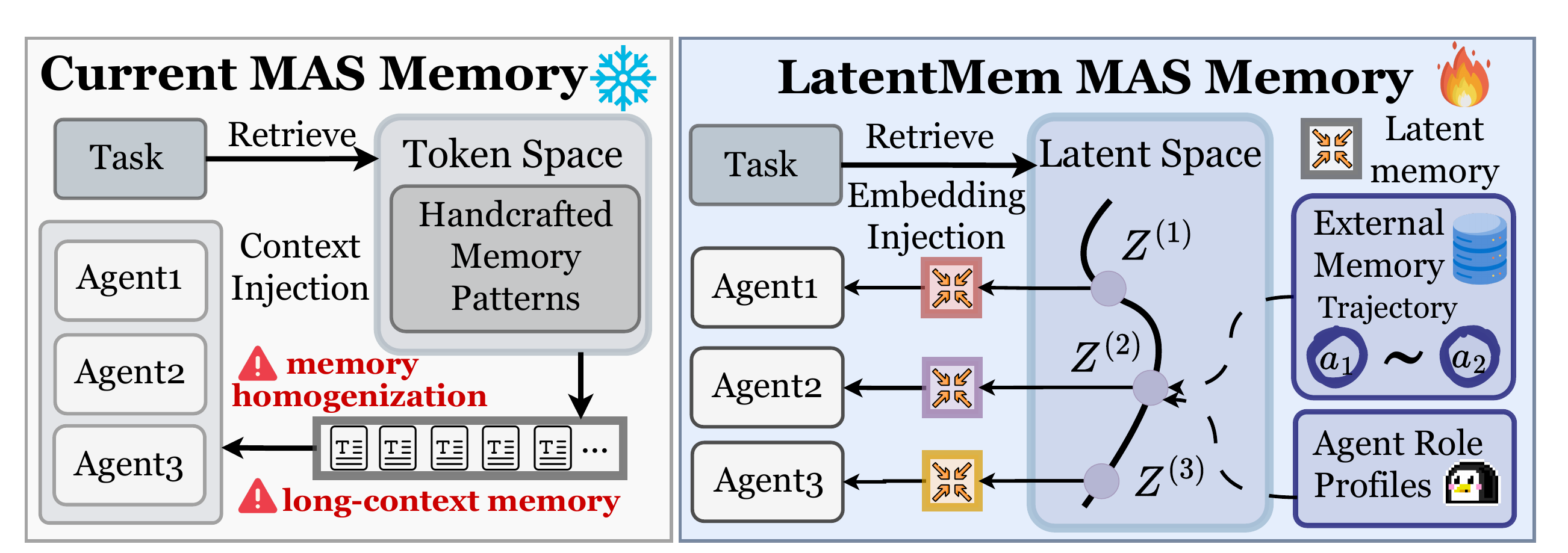}
  \caption{The paradigm comparison between existing multi-agent memory and \ourmethod. Instead of relying on handcrafted memory units, \ourmethod extracts agent-specific memories from the latent space by combining raw trajectories with agent profiles.}
  \label{fig:motivation}
\end{wrapfigure}
Large Language Model (LLM)-powered multi-agent systems (MAS), 
have emerged as a powerful framework for solving complex tasks by allowing agents to collaborate~\citep{ye2025x, zhang2024cut, yue2025masrouter} or compete~\citep{zhang2024can, yang2025revisiting}
beyond the capabilities of individual LLM agents. 
Pivotal to this success is the concept of multi-agent memory~\citep{wu2025human, hu2025memory}, which enables agents to accumulate, retain, and reuse experiences through interactions with both other agents and the environment, thereby supporting more coherent coordination and continual adaptation.

Building on this memory foundation, recent studies have increasingly explored multi-granularity memory repositories that capture experiences at different levels of abstraction, including \textit{(i) MAS trajectories}~\citep{qian2024scaling, wang2025mirix}, \textit{(ii) distilled semantic insights}~\citep{liu2025joyagent, zhu2025oagents}, and \textit{(iii) orchestrable skill schemas}~\citep{han2025legomem, zhang2025memevolve}. These designs endow memory systems with the ability to capture diverse memory patterns, such as trajectory summarization and high-level insight extraction, enabling MAS to adaptively integrate past experiences and jointly refine decision-making strategies~\citep{zheng2026lifelong, tomilin2025meal}.

However, despite the growing sophistication of existing memory systems, they remain constrained by two key limitations: 
\textit{(i) Memory homogenization:} Most methods adopt a one-size-fits-all strategy, ignoring the functional heterogeneity of agents, which undermines role adherence and amplifies correlated errors~\citep{cemri2025multi}, weakening system robustness and hindering long-term adaptation. 
\textit{(ii) Information overload:} MAS inherently involves long interaction contexts~\citep{zhang2024cut}, and multi-granularity memory designs further amplify this burden by introducing large volumes of stored entries~\citep{wang2025mirix, zhang2025g}, ultimately overwhelming agents and obscuring critical decision signals. Given the aforementioned challenges, a natural question arises: 
\obsbox{\textit{Given long and complex contexts in MAS, can we design a learnable memory that is both role-aware and token-efficient, without extensive manual engineering?}}

To address these challenges, we propose \ourmethod, a latent multi-agent memory framework that materializes agent-aware memory customization via token-efficient latent memory generation. Specifically, \ourmethod consists of two components: a lightweight \textbf{experience bank} for storing and retrieving raw MAS trajectories, and a \textbf{memory composer} that leverages agent profiles to distill raw trajectories into role-aware, compact latent memories and integrate them into the agents’ reasoning process. To encourage the memory composer to distill transferable, high-utility latent representations from raw trajectories, we propose \textbf{L}atent \textbf{M}emory \textbf{P}olicy \textbf{O}ptimization (\textbf{LMPO}), which computes advantages from relative rewards within multi-agent rollouts, optimizes token-level objectives, and exploits latent memory differentiability to enable gradient backpropagation through the memory composer.

As a novel attempt in latent MAS memory, \ourmethod offers three principal advantages: 
\textbf{(I)} It conditions the memory composer on agent role profiles to customize role-aware latent memories, thereby mitigating \textit{memory homogenization}; 
\textbf{(II)} It encodes multi-agent memory as fixed-length latent representations rather than unbounded discrete textual traces, thereby mitigating \textit{information overload};
\textbf{(III)} It exploits LMPO and latent memory differentiability to enable autonomous \textit{memory internalization and reconstruction}, thereby avoiding language constraints and obviating the need for meticulously engineered memory architectures.

Extensive experiments across six benchmarks and four mainstream MAS frameworks demonstrate that \ourmethod achieves: 
\textbf{(I) high performance}, improving state-of-the-art MAS by up to $16.20\%$ and $18.45\%$ in knowledge QA and code generation tasks, respectively; 
\textbf{(II) high efficiency}, using $50\%$ fewer tokens and reducing inference time to $\sim2/3$ compared to mainstream memory designs; and
\textbf{(III) strong generalization}, with out-of-domain datasets such as PDDL showing a $7.10\%$ improvement, and unseen MAS such as CAMEL exhibiting a $7.90\%$ gain compared to the vanilla setting. These results establish \ourmethod as a novel and effective framework for MAS memory.





\vspace{-0.5em}
\section{Related Works}

\textbf{LLM-Based Multi-Agent Systems.} MAS is a framework in which multiple agents collaborate by assuming distinct roles to achieve shared objectives~~\citep{guo2024large,li2024survey,tran2025multi}. Our work focuses on leveraging MAS for task-specific problem solving. Early task-solving MAS frameworks~\citep{hong2023metagpt,li2023camel,wu2024autogen,du2023improving,liang2023encouraging,zhang2024chain,wang2024moa} typically rely on predefined, static workflows, while more recent studies~\citep{zhuge2024gptswarm,zhang2024g,yang2025agentnet,zhou2025reso,zhou2025multi,zhang2025multi,yue2025masrouter,yang2026toward} have enabled MAS to dynamically reconfigure their organizational structures, improving adaptability to diverse and complex tasks while reducing computational costs. However, these methods typically require extensive searches over the design space, resulting in considerable computational and token overhead. Consequently, developing a lightweight mechanism for flexible MAS adaptation remains an open and challenging research problem.

\textbf{Memory in Multi-Agent Systems.} Memory enables agents to accumulate experience through interactions, thereby supporting coherent coordination and continual adaptation~\citep{xu2025mem,hu2025memory}. It plays a crucial role in task-solving and social simulation; our focus lies primarily on the former. Early memory designs in MAS typically rely on simple, within-trial mechanisms coupled to the system itself, such as a shared pool storing raw trajectories~\citep{hong2023metagpt,chen2023agentverse,yin2023exchange,qian2023communicative,qian2024chatdev,rezazadeh2025collaborative}. Modern memories, by contrast, have shifted towards more intricate and flexible structures. Representative examples include OAgents, which employs multi-granularity memory~\citep{zhu2025oagents}; EvolveR~\citep{wu2025evolver} and Agent KB~\citep{tang2025agent}, which compress raw trajectories into high-level semantic units; and MIRIX~\citep{wang2025mirix}, which transforms user goals into orchestrable procedural memories. However, these approaches overlook heterogeneous, role-aware memory design. \ourmethod addresses this limitation by equipping each agent with a compact, role-aware latent memory, thereby reinforcing role compliance, enhancing coordination, and improving continual adaptation.

\begin{figure*}[t]
    \centering
    \includegraphics[width=\linewidth]{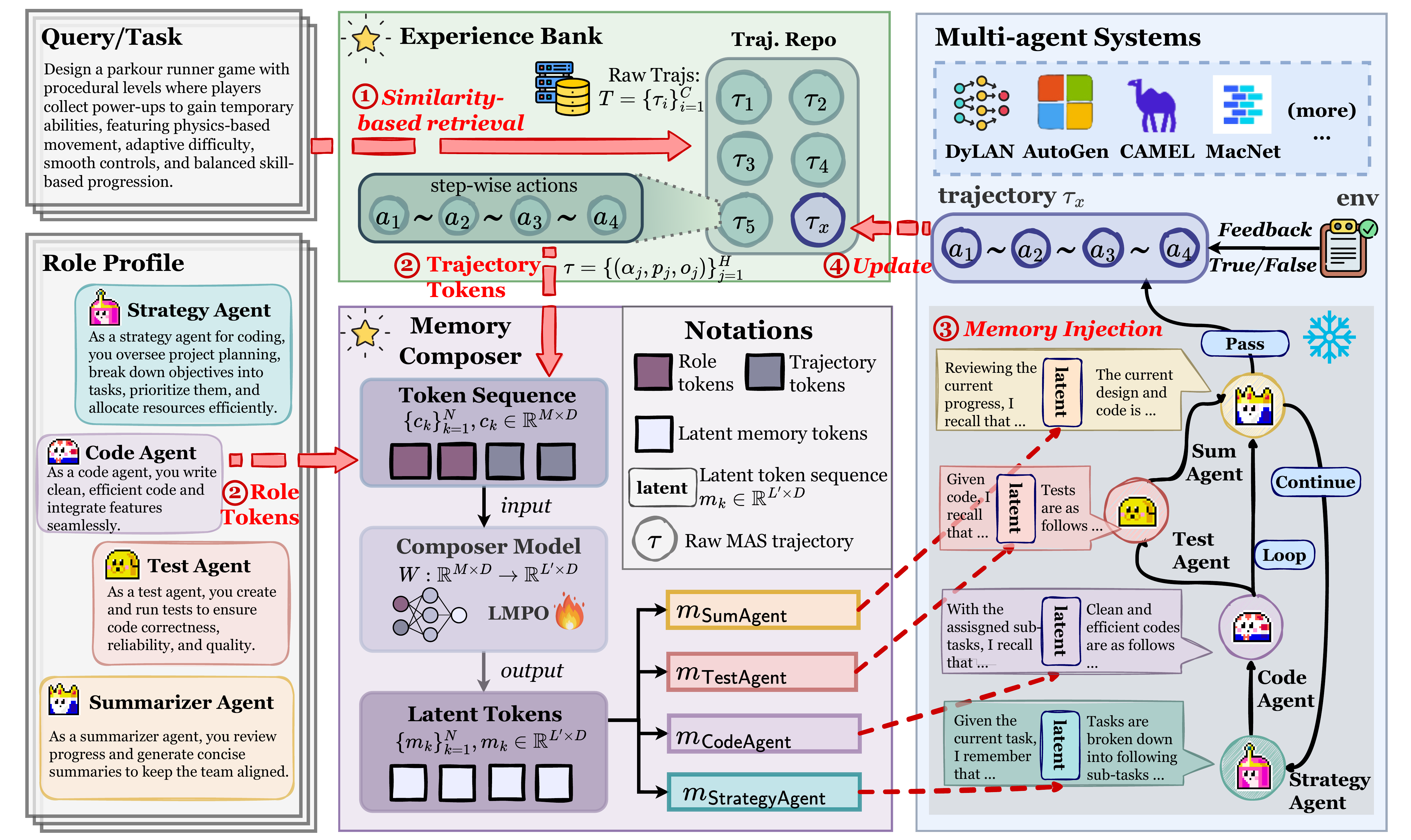}
    \caption{Overview of \ourmethod. The framework proceeds as follows: (1) retrieve relevant trajectories from the experience bank; (2) compress them with agent role profiles into latent memories via the LMPO-trained memory composer; (3) inject these memories into agent reasoning processes without altering the agent architectures; and (4) store new trajectories for continual improvement.}
    \vspace{-1em}
    \label{fig:method}
\end{figure*}

\section{Preliminary}\label{sec:prelim}
\paragraph{Notations.} 
Consider a multi-agent system $\mathcal{X}$ containing $N$ agents $\mathcal{A} = \{a_1, a_2, \dots, a_N\}$, and equipped with a global memory module $\mathcal{M}$ that stores and retrieves shared information among agents. Formally, the system can be represented as the tuple:
\begin{equation}
\mathcal{X} = (\mathcal{A}, \mathcal{G}, \mathcal{M})
\end{equation}
Each agent $a_k = (\gamma_k, \pi_{\theta_k})$ is defined by a role profile $\gamma_k$ and a policy $\pi_{\theta_k}$ parameterized by $\theta_k$. During execution, an agent receives an input prompt $p$ and a retrieved memory $m$ from $\mathcal{M}$, and produces a response $o$, denoted as $o = a_k(p, m)$. The execution graph $\mathcal{G}$ governs the topological order in which agents operate. Depending on the system architecture, $\mathcal{G}$ can be instantiated as either a static predefined topology~\citep{qian2024scaling} or a centralized dynamic regulation mechanism~\citep{wang2025talk}.

\paragraph{Problem Formulation.} 
Our objective is to find the memory module $\mathcal{M}$ that maximizes the expected performance of MAS $\mathcal{X}$, which is formally defined as: 
\begin{equation}
\max_{\mathcal{M}} \mathbb{E}_{q \sim \mathcal{D}, \tau \sim \mathcal{X}(q)} \left[ R(\tau) \right]
\end{equation}
where $\mathcal{D}$ denotes the dataset and $q$ is a query sampled from it. The system $\mathcal{X}$ processes the query $q$ to produce a reasoning trajectory $\tau$, and the reward function $R$ extracts the final answer from $\tau$ and evaluates its correctness. 

This formulation is agnostic to specific memory architectures, ranging from hand-crafted symbolic systems to learnable, parameterized counterparts. Conventional memory systems often rely on predefined patterns to accumulate experiences, while our approach  adopts a learnable memory module that generates compact, role-aware latent representations for dynamic integration into each agent's reasoning.

\section{Methodology}

In this section, we first present the overall pipeline of our proposed \ourmethod framework ($\triangleright$ \Cref{sec:pipeline}). Then we detail each module within the framework, including the experience bank ($\triangleright$ \Cref{sec:bank}) and the memory composer ($\triangleright$ \Cref{sec:composer}). Finally, we introduce Latent Memory Policy Optimization (LMPO), which enables end-to-end optimization of the memory composer through task-level feedback and reinforcement learning algorithm ($\triangleright$ \Cref{sec:lmpo}).

\subsection{Overall Pipeline}\label{sec:pipeline}
The overall pipeline of our proposed \ourmethod framework is illustrated in ~\Cref{fig:method}. 
It consists of two core components: a lightweight experience bank $\mathcal{B}$ for storing and retrieving historical trajectories, and a learnable memory composer $\mathcal{C}$ that transforms the retrieved relevant trajectories into compact, role-aware latent memories.

Upon receiving a new query, \ourmethod first retrieves a subset of relevant trajectories from the experience bank. These trajectories capture the interactions and behaviors of agents in previous MAS executions, forming a historical context that can guide reasoning. The retrieved trajectories, together with each agent’s role profile, are then processed by the memory composer, which distills them into compact latent memories tailored to the corresponding agents. During each agent’s reasoning process, these latent memories are appended to the token embeddings as additional latent tokens, forming a memory-augmented representation that incentivizes the agent to leverage prior experience and generate improved outputs. After the MAS completes a task, the newly generated trajectory is appended to the experience bank, enabling incremental accumulation of experiences.

This procedure forms a self-improving loop, allowing \ourmethod to continuously refine agent reasoning, support long-horizon coordination, and enhance continual adaptation. Moreover, the injection of latent memories maintains end-to-end differentiability of the entire forward process, facilitating efficient RL-based post-training ~\citep{qu2025survey} without incurring the heavy computation of retraining foundation models.

\subsection{Experience Bank}\label{sec:bank}
To accurately record historical MAS trajectories for future reuse, we construct an extremely lightweight experience bank $\mathcal{B}$. In line with the principle that scalable systems should rely on general learning mechanisms rather than hand-crafted knowledge~\citep{sutton2019bitter}, this bank stores and retrieves only raw trajectories, without introducing any human priors such as trajectory condensation~\citep{wang2025mirix} or insight extraction~\citep{zhao2024expel},

\textbf{Initialization.}  
We populate the experience bank $\mathcal{B}$ with a wide-ranging collection of trajectories covering multiple domains and MAS frameworks to enable the memory composer $\mathcal{C}$ to learn generalizable memory patterns across diverse domains and agent coordination patterns. The resulting initialized bank is denoted as $\{\tau_i\}_{i=1}^C$, where $C$ specifies its initial capacity. Each trajectory $\tau = \{(\alpha_j, p_j, o_j)\}_{j=1}^{H}$ records, at each step, the index of the active agent $\alpha_j$ along with its input prompt $p_j$ and corresponding output $o_j$, where $H$ denotes the trajectory horizon.

\textbf{Retrieval.} Upon receiving a new user query $q$, \ourmethod performs similarity-based retrieval over $\mathcal{B}$ to obtain a subset of $K$ relevant trajectories $\mathcal{T}_q$:
\begin{equation}\label{eq:similarity}
\mathcal{T}_q = \text{top-}K_{\tau_i \in \mathcal{B}} \left( \text{sim}(\mathbf{v}(q), \mathbf{v}(\tau_i)) \right) = \{\tau_i\}_{i=1}^K,
\end{equation}
where $\mathbf{v}(\cdot)$ maps queries or trajectories into a latent embedding space, e.g., using MiniLM~\citep{wang2020microsoft}, and $\text{sim}(\cdot,\cdot)$ denotes the cosine similarity. The retrieved trajectories will be subsequently processed by the memory composer, which distills them into latent memories to guide subsequent MAS reasoning tasks.

\textbf{Update.} Once a task is completed, the new trajectory $\tau_{\text{new}}$ is appended to the experience bank $\mathcal{B}$ for future reuse:
\begin{equation}\label{eq:exp_update}
\mathcal{B} \leftarrow \mathcal{B} \cup \{\tau_{\text{new}}\}.
\end{equation}
This streamlined update mechanism allows \ourmethod to incrementally accumulate experiences online during inference, facilitating continual adaptation and cross-task coordination without the need for retraining.

However, directly feeding the retrieved raw trajectories to agents is suboptimal, as it can overwhelm LLMs with excessive context~\citep{cemri2025multi} and fails to capture role-specific representations in heterogeneous MAS~\citep{subramaniam2025multiagent}. To address these limitations, we introduce the memory composer $\mathcal{C}$, which effectively transforms low-level raw trajectories into compact, high-level, role-aware latent memories.

\subsection{Memory Composer}\label{sec:composer}
After identifying the relevant raw trajectories $\mathcal{T}_q$, we introduce the memory composer $\mathcal{C}$, which provides each agent with generalizable memories. 
Formally, $\mathcal{C}$ is instantiated as a deep neural network $\sigma_\phi$ parameterized by $\phi$. At each reasoning step $j$, $\sigma_\phi$ takes as input the retrieved trajectories $\mathcal{T}_q$ and the role profile $\gamma_{\alpha_j}$ of the active agent $a_{\alpha_j}$, producing a fixed-length, agent-aware latent memory matrix:
\begin{equation}\label{eq:composer}
m_j = \sigma_\phi(\gamma_{\alpha_j}, \mathcal{T}_q) \in \mathbb{R}^{L' \times D},
\end{equation}
where $L'$ is a fixed length of the latent memory and $D$ denotes the hidden dimension of the foundation model.

To conduct reasoning, the active agent $a_{\alpha_j}$ first encodes its input prompt $p_j$ into a sequence of hidden state vectors $h_j = (h_j^{(1)}, \dots, h_j^{(L)}) \in \mathbb{R}^{L \times D}$. The agent’s latent memory $m_{j}$ is then concatenated to $h_{j}$ to form an extended input shaped $\mathbb{R}^{(L+L') \times D}$, resulting in a memory-augmented policy:
\begin{equation}
\tilde{\pi}_{\theta_{\alpha_j}}(p_{j}, m_{j}) = \pi_{\theta_{\alpha_j}}(\text{concat}(h_{j}, m_{j})),
\end{equation}
where $\tilde{\pi}_{\theta_{\alpha_j}}$ is a wrapped version of $\pi_{\theta_{\alpha_j}}$ that seamlessly incorporates memory injection at the model level, remaining transparent to the agent layer and requiring no modifications to the system architecture.

\subsection{Latent Memory Policy Optimization (LMPO)}\label{sec:lmpo}
To enable end-to-end optimization of \ourmethod while preserving strong generalization across diverse domains and MAS frameworks, we propose Latent Memory Policy Optimization (LMPO), a variant of GRPO~\citep{shao2024deepseekmath}, which encourages the memory composer to generate transferable, high-utility latent representations.

\paragraph{Parametric Dependency.} 
We first describe the gradient flow during LMPO, in which the learning signal propagates through the latent memories to optimize the memory composer $\mathcal{C}$, while keeping the agent backbones $\{\theta_k\}_{k=1}^N$ frozen. Formally, given a query $q$ and the retrieved trajectories $\mathcal{T}_q$ from the experience bank $\mathcal{B}$, the generation of a new trajectory $\tau = \{(\alpha_j, p_j, o_j)\}_{j=1}^H$ is factorized sequentially as:
\begin{equation}
\mathbb{P}(\tau \mid q, \mathcal{T}_q; \phi, \{\theta_k\}_{k=1}^N) = \prod_{j=1}^{H} \mathbb{P}(o_{j} \mid p_{j}, m_j; \theta_{\alpha_j}).
\end{equation}
Crucially, the latent memory $m_j = \sigma_\phi(\mathcal{T}_q, \gamma_{\alpha_j})$, as defined in \Cref{eq:composer}, 
serves as a differentiable interface through which $\phi$ influences the autoregressively generated output $o_j$ of the active agent $a_{\alpha_j}$ at reasoning step $j$:
\begin{equation}
\mathbb{P}(o_j \mid p_j, m_j; \theta_{\alpha_j}) = \prod_{t=1}^{T} \tilde{\pi}_{\theta_{\alpha_j}}( o_j^{(t)} \mid p_j, o_j^{(<t)}, m_j).
\end{equation}
Since the composite policy $\tilde{\pi}_{\theta_{\alpha_j}}$ is conditioned on $m_j$, the gradient of any task-level objective can be backpropagated through the agent’s forward pass to refine $\phi$. This dependency ensures that the memory composer can be optimized end-to-end to produce high-quality latent memories, forming the basis of our policy optimization strategy.

\paragraph{Policy Optimization.}
Building on the differentiable path above, LMPO leverages task-level feedback through latent memories as a bridge to directly optimize memory composer attached to the MAS, encouraging it to distill high-utility, agent-specific memories from retrieved raw trajectories and thereby enhance reasoning quality and overall performance.

Formally, given a query $q$ and its retrieved relevant trajectories $\mathcal{T}_q$, we sample a group of $G$ trajectories:
\begin{equation}
\{\hat{\tau}_i\}_{i=1}^G \sim \mathbb{P}(\cdot \mid q, \mathcal{T}_q; \phi, \{ \theta_k \}_{k=1}^N).
\end{equation}
Each trajectory is evaluated using reward $R(\hat{\tau}_i)$, and its relative quality is captured by the group-based advantage:
\begin{equation}
\hat{A}_i = \frac{R(\hat{\tau}_i) - \text{mean}(\{ R(\hat{\tau}_i) \}_{i=1}^G)}{\text{std}(\{ R(\hat{\tau}_i) \}_{i=1}^G) + \epsilon}.
\end{equation}

While standard reinforcement learning ~\citep{zhang2025survey} often employs trajectory-level objectives, such approaches treat all sequences equally, causing tokens in longer MAS interactions to contribute disproportionately less to the gradient~\citep{yu2025dapo}. This makes it difficult for the memory composer to capture critical coordination patterns within long-horizon tasks. Instead, we adopt a token-level surrogate objective:
\begin{equation}
\begin{split}
\mathcal{J}_{\text{LMPO}}(\phi) &= \mathbb{E}_{q \sim \mathcal{D}, \{ \hat{\tau}_i \}_{i=1}^G \sim \mathbb{P}(\cdot \mid q, \mathcal{T}_q)} \\
&\quad \Biggl[\frac{1}{|\{\hat\tau_i\}_{i=1}^G|}\sum_{i,j,t} \mathcal{L}_{i,j,t}(\phi) \Biggr],
\end{split}
\end{equation}
where $|\{\hat\tau_i\}_{i=1}^G|$ is the total number of generated tokens within the trajectory group and $\mathcal{L}_{i,j,t}(\phi)$ is defined as:
\begin{equation}
\begin{split}
\mathcal{L}_{i,j,t}(\phi) &= \min \Big( r_{i,j,t}(\phi)\, \hat{A}_{i}, \\
& \quad \quad \text{clip}\big( r_{i,j,t}(\phi), 1-\varepsilon, 1+\varepsilon \big)\, \hat{A}_{i} \Big),
\end{split}
\end{equation}
and the token-level importance sampling ratio
\begin{equation}
r_{i,j,t}(\phi) = \frac{\tilde{\pi}_{\theta}(o_{i,j}^{(t)} \mid p_{i,j}, o_{i,j}^{(<t)}, \sigma_\phi(\gamma_{\alpha_{i,j}}, \mathcal{T}_q))}{\tilde{\pi}_{\theta}(o_{i,j}^{(t)} \mid p_{i,j}, o_{i,j}^{(<t)}, \sigma_{\phi_{\text{old}}}(\gamma_{\alpha_{i,j}}, \mathcal{T}_q))}
\end{equation}
measures how the policy of agent $a_{\alpha_{i,j}}$ at reasoning step $j$ and token $t$ is modulated by the updated memory.

\begin{table*}[!t]
\caption{Performance comparison across diverse memory frameworks on six benchmarks, using \llmname{Qwen3-4B-Instruct-2507} as the backbone. The \hlfirst{best} and \hlsecond{second best} results are highlighted. TriviaQA, KodCode, StrategyQA, and PopQA are in-domain benchmarks, while BigCodeBench and PDDL are out-of-domain. AutoGen and MacNet are in-distribution MAS frameworks, whereas CAMEL and DyLAN are unseen frameworks introducing new agent roles and collaboration patterns. Additional details are provided in Appendix \ref{app:train_details}.
}
\label{tab:results_4b}

\renewcommand{\arraystretch}{1.1}
\centering \scriptsize \setlength{\tabcolsep}{8pt}

\resizebox{\textwidth}{!}{
\begin{tabular}{rl|l|llll|ll|l}
\toprule
\multicolumn{2}{c|}{\multirow{2}{*}{}} & \multicolumn{1}{c|}{\multirow{2}{*}{\textbf{Method}}} & \multicolumn{4}{c|}{\textbf{Held-in}} & \multicolumn{2}{c|}{\textbf{Held-out}} & \multirow{2}{*}{\textbf{Average}} \\
\multicolumn{2}{c|}{} & & \multicolumn{1}{c}{\textbf{TriviaQA}} & \multicolumn{1}{c}{\textbf{KodCode}}  & \multicolumn{1}{c}{\textbf{StrategyQA}} & \multicolumn{1}{c|}{\textbf{PopQA}} & \multicolumn{1}{c}{\textbf{BigCodeBench}} & \multicolumn{1}{c|}{\textbf{PDDL}} \\
\midrule

\multirow{18}{*}{\rotatebox[origin=c]{90}{\textbf{Held-in}}\hspace{-3em}} & \multirow{9}{*}{\rotatebox[origin=c]{90}{\textbf{AutoGen}}} & No-memory & 60.31 & 68.40 & 58.25 & 38.78 & 79.53 & 16.39 & 53.61 \\
& & ChatDev & 57.34\degrade{2.97} & 68.55\upgrade{0.15} & 59.18\upgrade{0.93} & 33.24\degrade{5.54} & 79.15\degrade{0.38} & 15.22\degrade{1.17} & 52.11\degrade{1.50} \\
& & MetaGPT & 60.35\upgrade{0.04} & 70.05\upgrade{1.65} & 60.92\upgrade{2.67} & 33.80\degrade{4.98} & 80.40\upgrade{0.87} & 11.95\degrade{4.44} & 52.91\degrade{0.70} \\
& & Generative & 59.65\degrade{0.66} & 70.90\upgrade{2.50} & 62.66\upgrade{4.41} & 40.37\upgrade{1.59} & 81.14\upgrade{1.61} & 13.94\degrade{2.45} & 54.78\upgrade{1.17} \\
& & Voyager & 57.50\degrade{2.81} & 68.95\upgrade{0.55} & 60.48\upgrade{2.23} & 33.56\degrade{5.22} & 79.40\degrade{0.13} & 13.62\degrade{2.77} & 52.25\degrade{1.36} \\
& & G-Memory & \hlsecond{60.56\upgrade{0.25}} & \hlsecond{71.40\upgrade{3.00}} & \hlsecond{63.89\upgrade{5.64}} & \hlsecond{42.67\upgrade{3.89}} & \hlsecond{80.96\upgrade{1.43}} & \hlsecond{17.06\upgrade{0.67}} & \hlsecond{56.09\upgrade{2.48}} \\
& & JoyAgent & 59.44\degrade{0.87} & 70.90\upgrade{2.50} & 62.33\upgrade{4.08} & 41.89\upgrade{3.11} & 80.52\upgrade{0.99} & 14.26\degrade{2.13} & 54.89\upgrade{1.28} \\
& & OAgents & 59.85\degrade{0.46} & 70.80\upgrade{2.40} & 62.45\upgrade{4.20} & 40.70\upgrade{1.92} & 81.32\upgrade{1.79} & 16.70\upgrade{0.31} & 55.30\upgrade{1.69} \\
& & \ourmethod & \hlfirst{76.51\upgrade{16.20}} & \hlfirst{76.80\upgrade{8.40}} & \hlfirst{65.48\upgrade{7.23}} & \hlfirst{52.70\upgrade{13.92}} & \hlfirst{81.49\upgrade{1.96}} & \hlfirst{23.49\upgrade{7.10}} & \hlfirst{62.75\upgrade{9.14}} \\
\cmidrule{3-10}

& \multirow{9}{*}{\rotatebox[origin=c]{90}{\textbf{MacNet}}} & No-memory & 53.77 & 70.40 & 56.44 & 24.89 & 78.15 & 20.73 & 50.73 \\
& & ChatDev & 57.29\upgrade{3.52} & 70.50\upgrade{0.10} & 60.22\upgrade{3.78} & 35.33\upgrade{10.44} & 79.17\upgrade{1.02} & 15.85\degrade{4.88} & 53.06\upgrade{2.33} \\
& & MetaGPT & 61.16\upgrade{7.39} & 71.50\upgrade{1.10} & 61.14\upgrade{4.70} & 33.98\upgrade{9.09} & 78.33\upgrade{0.18} & 17.81\degrade{2.92} & 53.99\upgrade{3.26} \\
& & Generative & 59.89\upgrade{6.12} & 71.15\upgrade{0.75} & 61.79\upgrade{5.35} & 43.39\upgrade{18.50} & 79.47\upgrade{1.32} & 16.81\degrade{3.92} & 55.42\upgrade{4.69} \\
& & Voyager & 58.19\upgrade{4.42} & 69.80\degrade{0.60} & 61.35\upgrade{4.91} & 35.38\upgrade{10.49} & \hlsecond{80.53\upgrade{2.38}} & 14.75\degrade{5.98} & 53.33\upgrade{2.60} \\
& & G-Memory & \hlsecond{62.43\upgrade{8.66}} & \hlsecond{72.50\upgrade{2.10}} & 62.20\upgrade{5.76} & \hlsecond{43.88\upgrade{18.99}} & 80.44\upgrade{2.29} & 21.82\upgrade{1.09} & \hlsecond{57.21\upgrade{6.48}} \\
& & JoyAgent & 61.33\upgrade{7.56} & 70.80\upgrade{0.40} & \hlsecond{62.34\upgrade{5.90}} & 43.22\upgrade{18.33} & 79.80\upgrade{1.65} & 21.20\upgrade{0.47} & 56.45\upgrade{5.72} \\
& & OAgents & 60.63\upgrade{6.86} & 71.30\upgrade{0.90} & 61.79\upgrade{5.35} & 41.90\upgrade{17.01} & 78.86\upgrade{0.71} & \hlsecond{22.83\upgrade{2.10}} & 56.22\upgrade{5.49} \\
& & \ourmethod & \hlfirst{65.98\upgrade{12.21}} & \hlfirst{78.90\upgrade{8.50}} & \hlfirst{64.46\upgrade{8.02}} & \hlfirst{44.14\upgrade{19.25}} & \hlfirst{81.49\upgrade{3.34}} & \hlfirst{25.13\upgrade{4.40}} & \hlfirst{60.02\upgrade{9.29}} \\
\midrule

\multirow{18}{*}{\rotatebox[origin=c]{90}{\textbf{Held-out}}\hspace{-3em}} & \multirow{9}{*}{\rotatebox[origin=c]{90}{\textbf{CAMEL}}} & No-memory & 56.96 & 70.70 & 58.04 & 32.38 & 79.14 & 22.10 & 53.22 \\
& & ChatDev & 57.55\upgrade{0.59} & 68.20\degrade{2.50} & 59.04\upgrade{1.00} & 36.78\upgrade{4.40} & 78.12\degrade{1.02} & 18.58\degrade{3.52} & 53.04\degrade{0.18} \\
& & MetaGPT & 59.06\upgrade{2.10} & 69.90\degrade{0.80} & 59.38\upgrade{1.34} & 37.47\upgrade{5.09} & \hlsecond{80.53\upgrade{1.39}} & 22.55\upgrade{0.45} & 54.82\upgrade{1.60} \\
& & Generative & 57.63\upgrade{0.67} & 70.65\degrade{0.05} & \hlsecond{61.20\upgrade{3.16}} & 35.32\upgrade{2.94} & 80.35\upgrade{1.21} & 17.29\degrade{4.81} & 53.74\upgrade{0.52} \\
& & Voyager & 56.57\degrade{0.39} & 69.85\degrade{0.85} & 58.27\upgrade{0.23} & 36.25\upgrade{3.87} & 79.91\upgrade{0.77} & 23.65\upgrade{1.55} & 54.08\upgrade{0.86} \\
& & G-Memory & \hlsecond{59.20\upgrade{2.24}} & 70.40\degrade{0.30} & 60.12\upgrade{2.08} & \hlsecond{38.04\upgrade{5.66}} & 79.28\upgrade{0.14} & \hlsecond{24.56\upgrade{2.46}} & \hlsecond{55.27\upgrade{2.05}} \\
& & JoyAgent & 58.10\upgrade{1.14} & 70.20\degrade{0.50} & 61.34\upgrade{3.30} & 37.50\upgrade{5.12} & 79.45\upgrade{0.31} & 20.65\degrade{1.45} & 54.54\upgrade{1.32} \\
& & OAgents & 58.33\upgrade{1.37} & \hlsecond{71.40\upgrade{0.70}} & 60.09\upgrade{2.05} & 31.99\degrade{0.39} & 79.28\upgrade{0.14} & 13.89\degrade{8.21} & 52.50\degrade{0.72} \\
& & \ourmethod & \hlfirst{68.74\upgrade{11.78}} & \hlfirst{77.75\upgrade{7.05}} & \hlfirst{64.20\upgrade{6.16}} & \hlfirst{47.23\upgrade{14.85}} & \hlfirst{80.70\upgrade{1.56}} & \hlfirst{28.12\upgrade{6.02}} & \hlfirst{61.12\upgrade{7.90}} \\
\cmidrule{3-10}

& \multirow{9}{*}{\textbf{\rotatebox[origin=c]{90}{\textbf{DyLAN}}}} & No-memory & 53.87 & 69.25 & 57.44 & 24.89 & 79.81 & 23.81 & 51.51 \\
& & ChatDev & 56.28\upgrade{2.41} & 68.55\degrade{0.70} & 60.23\upgrade{2.79} & 39.27\upgrade{14.38} & 79.82\upgrade{0.01} & 16.55\degrade{7.26} & 53.45\upgrade{1.94} \\
& & MetaGPT & \hlsecond{61.78\upgrade{7.91}} & 71.50\upgrade{2.25} & 62.23\upgrade{4.79} & 37.00\upgrade{12.11} & 78.25\degrade{1.56} & 19.19\degrade{4.62} & 54.99\upgrade{3.48} \\
& & Generative & 60.39\upgrade{6.52} & \hlsecond{71.70\upgrade{2.45}} & 60.92\upgrade{3.48} & \hlsecond{43.57\upgrade{18.68}} & 78.86\degrade{0.95} & 18.58\degrade{5.23} & 55.67\upgrade{4.16} \\
& & Voyager & 59.06\upgrade{5.19} & 70.05\upgrade{0.80} & 61.79\upgrade{4.35} & 40.40\upgrade{15.51} & 80.00\upgrade{0.19} & \hlsecond{26.37\upgrade{2.56}} & 56.28\upgrade{4.77} \\
& & G-Memory & 61.23\upgrade{7.36} & 71.35\upgrade{2.10} & \hlsecond{63.76\upgrade{6.32}} & 42.28\upgrade{17.39} & \hlsecond{80.35\upgrade{0.54}} & 17.94\degrade{5.87} & 56.15\upgrade{4.64} \\
& & JoyAgent & 59.96\upgrade{6.09} & 71.35\upgrade{2.10} & 62.12\upgrade{4.68} & 41.67\upgrade{16.78} & 79.89\upgrade{0.08} & 25.81\upgrade{2.00} & \hlsecond{56.80\upgrade{5.29}} \\
& & OAgents & 60.98\upgrade{7.11} & 71.15\upgrade{1.90} & 61.35\upgrade{3.91} & 42.36\upgrade{17.47} & 78.77\degrade{1.04} & 19.75\degrade{4.06} & 55.73\upgrade{4.21} \\
& & \ourmethod & \hlfirst{65.55\upgrade{11.68}} & \hlfirst{78.80\upgrade{9.55}} & \hlfirst{65.22\upgrade{7.78}} & \hlfirst{44.25\upgrade{19.36}} & \hlfirst{81.40\upgrade{1.59}} & \hlfirst{29.08\upgrade{5.27}} & \hlfirst{60.72\upgrade{9.21}} \\
\bottomrule

\end{tabular}
}
\vspace{-1em}

\end{table*}

\section{Experiments}\label{sec:exp}

\subsection{Experimental Setup}
\textbf{Datasets and Benchmarks.} Our evaluation covers six benchmarks across four domains: \textbf{(1) Knowledge-intensive QA}: TriviaQA~\citep{joshi-etal-2017-triviaqa} and PopQA~\citep{mallen2023not}; \textbf{(2) Coding}: KodCode~\citep{xu2025kodcode} and BigCodeBench~\citep{jain2024livecodebench}; \textbf{(3) Reasoning QA}: StrategyQA~\citep{geva2021did}; and \textbf{(4) Symbolic Planning}: PDDL~\citep{silver2024generalized}. Detailed information for these benchmarks are provided in Appendix \ref{app:dataset}.

\textbf{Baselines.} Besides memory-free methods, we select three representative single-agent memory baselines, including Voyager~\citep{wang2023voyager}, Generative~\citep{park2023generative}, and JoyAgent~\citep{liu2025joyagent}, as well as four multi-agent memory baselines adopted from mainstream MAS frameworks: MetaGPT~\citep{hong2023metagpt}, ChatDev~\citep{qian2024chatdev}, OAgents~\citep{zhu2025oagents}, JoyAgent~\cite{liu2025joyagent}, and MAS-specific G-Memory~\citep{zhang2025g}. Additional details are provided in Appendix~\ref{app:baseline}.

\textbf{MAS and LLM Backbones.} Four representative multi-agent frameworks are adopted to integrate with \ourmethod and the baselines, including AutoGen~\citep{wu2024autogen}, MacNet~\citep{qian2024scaling}, CAMEL~\citep{li2023camel} and DyLAN~\citep{liu2024dynamic}. More details on the MAS setups are placed in Appendix \ref{app:mas}. 
To instantiate these MAS frameworks, we adopt two common LLMs with different sizes, i.e., \llmname{Qwen/Qwen3-4B-Instruct-2507} and \llmname{meta-llama/Llama-3.1-8B-Instruct}.

\textbf{Training Configurations.} We implement the embedding function $\mathbf{v}(\cdot)$ mentioned in \Cref{eq:similarity} with the \llmname{all-MiniLM-L6-v2} model~\cite{wang2020minilm}. The memory composer $\mathcal{C}$ is realized as a lightweight transformer, with its parameters initialized from the backbone LLM and trained using LoRA~\cite{hu2022lora}. We set $K=1$ in \Cref{eq:similarity} and fix the latent memory sequence length to $L' = 8$. The ablation study on hyper-parameter settings is reported in \Cref{sec:ablation}. Detailed training setups and parameter configurations are listed in Appendix \ref{app:train_details}.

\subsection{Main Results}
\textbf{\ourmethod Delivers High-Performance Memory Across Domains and MAS Frameworks.} 
As shown in \Cref{tab:results_4b}, when integrated with in-domain MAS frameworks such as AutoGen and MacNet that are powered by \llmname{Qwen3-4B-Instruct-2507}, \ourmethod outperforms the state-of-the-art single- and multi-agent memory baselines by an average of $7.86\%$ and $6.66\%$, respectively. Notably, it yields a $16.20\%$ improvement for AutoGen on the TriviaQA benchmark. Furthermore, \ourmethod exhibits strong scalability with the model size increases. 
As shown in Appendix \ref{tab:results_8b}, it elevates MacNet’s performance on KodCode from $48.50\%$ to $65.50\%$ using \llmname{Llama-3.1-8B-Instruct}.

\textbf{\ourmethod Exhibits Strong Generalization Capability.} 
On out-of-domain benchmarks, most MAS memory methods fail to generalize. As shown in \Cref{tab:results_4b}, \ourmethod improves AutoGen on PDDL by 7.10\%, while MetaGPT and Voyager drop by up to 4.44\% and 2.77\%, respectively. Similarly, on previously unseen MAS frameworks, \ourmethod boosts CAMEL on KodCode by 7.05\%, whereas nearly all baselines decline. We attribute these gaps to the rigid and homogeneous memory designs of existing methods, which limit adaptability and representational capacity. These results demonstrate \ourmethod’s robustness across domains, agent roles, and collaboration patterns, highlighting the importance of role-aware memory for generalizable MAS.

\subsection{Cost Analysis}
\begin{wrapfigure}{r}{0.5\textwidth}
    \vspace{-2.0em}
    \centering
    \includegraphics[width=0.5\columnwidth]{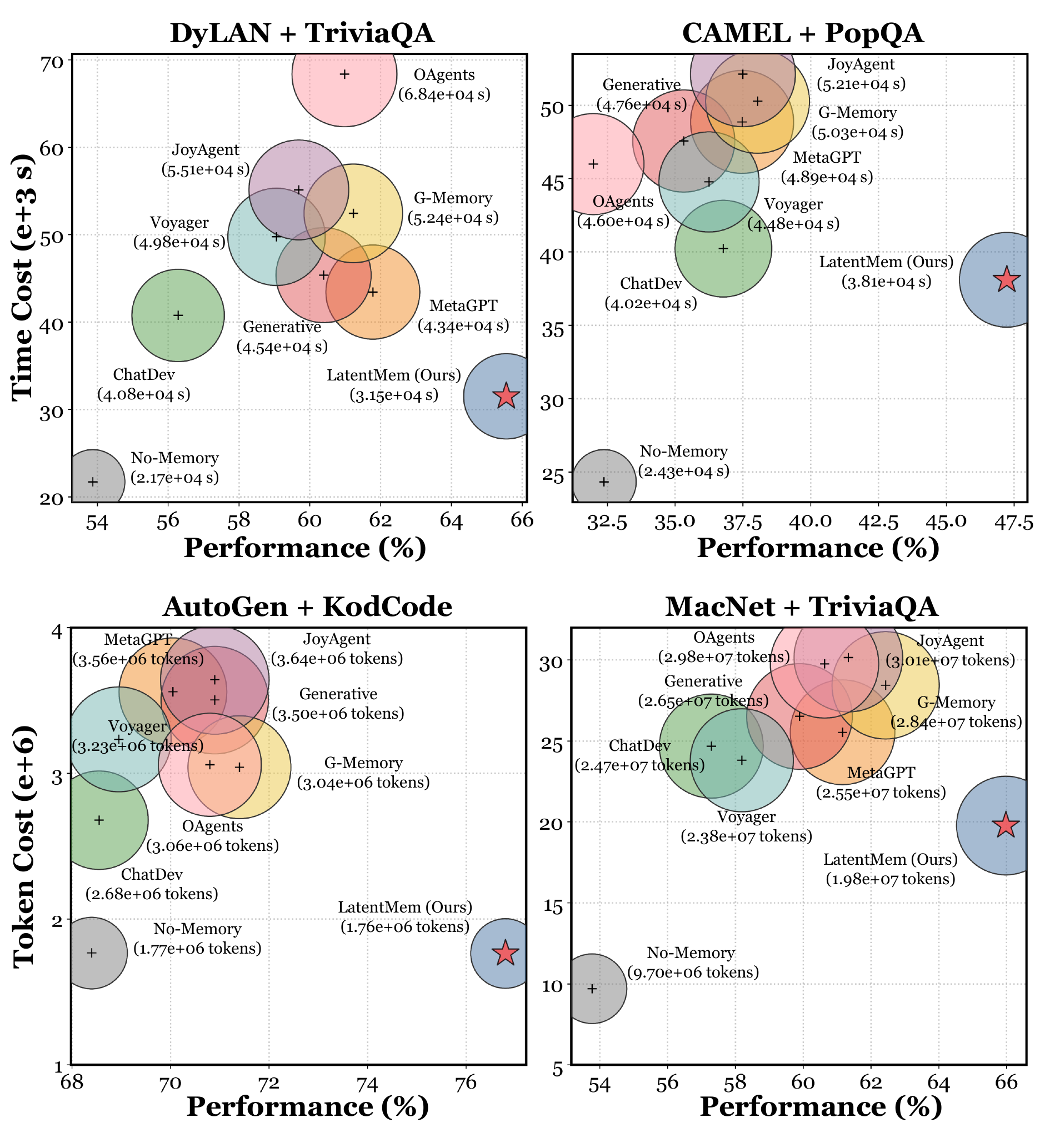}
    \caption{Time and token consumption of \ourmethod. Each panel shows the trade-off between performance and resource cost under different memory architectures: the top row plots performance versus time, the bottom row plots performance versus token cost. Circle area reflects relative resource consumption.}
    \vspace{-2.0em}
    \label{fig:cost}
\end{wrapfigure}

As shown in \Cref{fig:cost}, \ourmethod achieves the largest performance gains among memory-based baselines while incurring minimal time and token costs. It delivers the greatest improvement on TriviaQA for DyLAN ($+11.68\%$ over No-memory) with substantially lower time overhead (e.g., cutting inference time by a factor of $2.16\times$ relative to OAgents), and achieves the highest gain on KodCode for AutoGen ($+8.40\%$) while using even fewer tokens than No-Memory (0.01M tokens less). In contrast, JoyAgent consumes $1.87$M additional tokens for only a $2.50\%$ gain, highlighting the superior efficiency of \ourmethod.

\subsection{Comparison with Multi-Agent Fine-Tuning}
To assess the effectiveness of \ourmethod under comparable training conditions, we compare it with a representative multi-agent fine-tuning baseline, MARTI~\citep{marti2025}. MARTI is trained with GRPO under exactly the same computational budget as \ourmethod, where all agents share a single LLM backbone and are trained on the same datasets (TriviaQA, KodCode, StrategyQA, and PopQA), ensuring a fully fair and controlled comparison.

\begin{wraptable}{r}{0.5\textwidth}
\vspace{-1.0em}
\caption{Performance comparison between the multi-agent fine-tuning method MARTI and \ourmethod on KodCode and TriviaQA across two MAS frameworks, AutoGen and MacNet.}
\label{tab:marti}

\centering \scriptsize \setlength{\tabcolsep}{8pt}
\begin{tabular}{l|l|ll}
\toprule
\multicolumn{1}{c|}{\textbf{MAS}} & \multicolumn{1}{c|}{\textbf{Method}} & \multicolumn{1}{c}{\textbf{KodCode}} & \multicolumn{1}{c}{\textbf{TriviaQA}} \\
\midrule
\multirow{2}{*}{AutoGen} & MARTI & 74.20 & 64.78 \\
                         & \ourmethod & 76.80\upgrade{2.60} & 76.51\upgrade{11.73} \\
\midrule
\multirow{2}{*}{MacNet}  & MARTI & 73.10 & 62.31 \\
                         & \ourmethod & 78.90\upgrade{5.80} & 65.98\upgrade{3.67} \\
\bottomrule
\end{tabular}
\vspace{-1em}
\end{wraptable}

As shown in \Cref{tab:marti}, \ourmethod consistently outperforms direct agent backbone fine-tuning across all settings. Notably, on the TriviaQA dataset with the AutoGen framework, \ourmethod achieves a substantial improvement of $11.73\%$. Moreover, on more complex MAS settings such as MacNet, MARTI experiences a $1.10\%$ performance drop on KodCode compared to AutoGen, whereas \ourmethod instead surpasses its AutoGen counterpart by $2.10\%$. These results indicate that \ourmethod better exploits the structural advantages of complex MAS, leading to stronger performance gains than direct backbone fine-tuning.

\subsection{Framework Analysis}\label{sec:analysis}

\begin{wrapfigure}{r}{0.5\textwidth}
    \vspace{-1.0em}
    \centering
    \includegraphics[width=0.5\columnwidth]{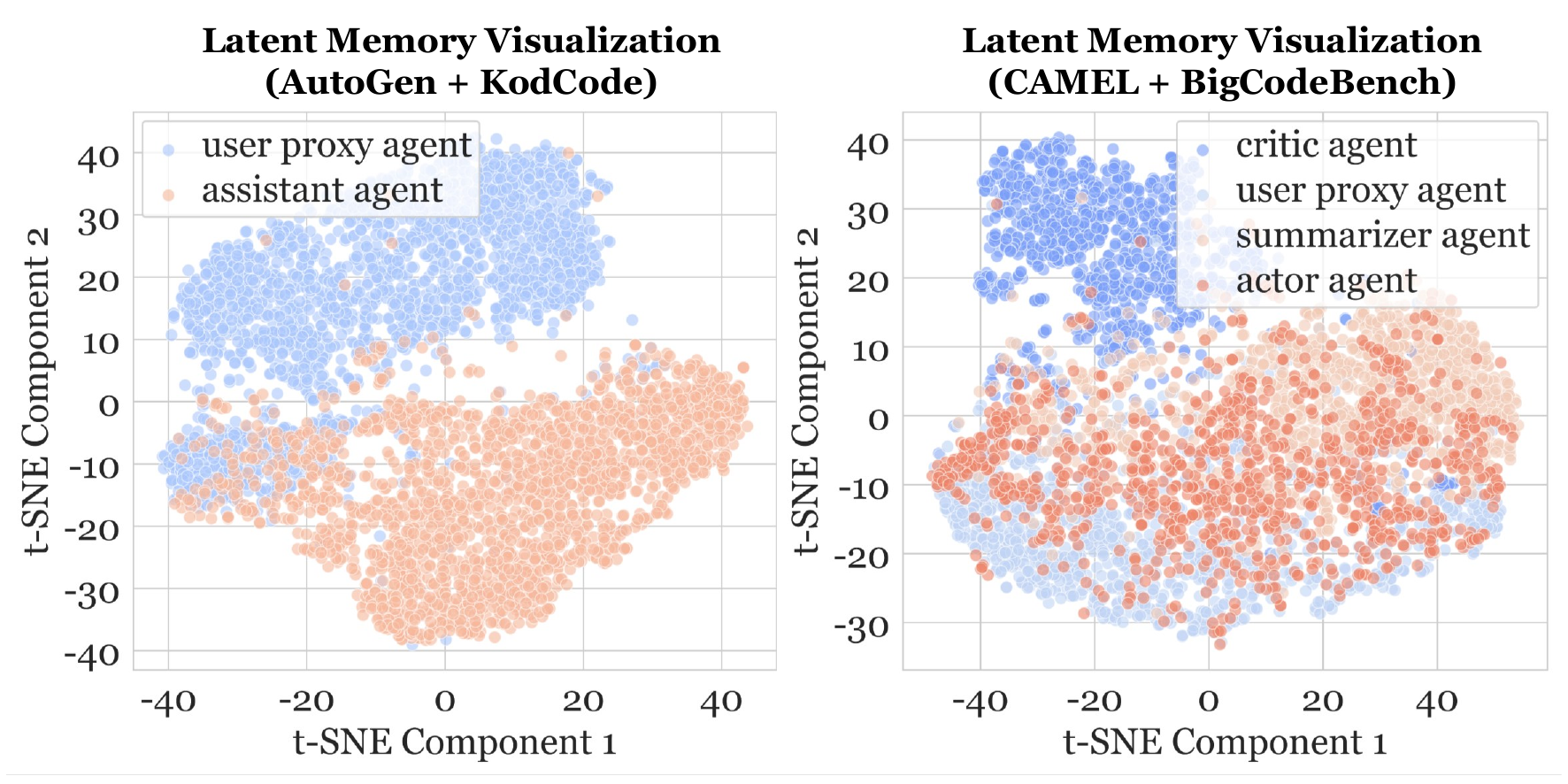}
    \caption{t-SNE visualization of latent memories generated by \ourmethod across different datasets and MAS frameworks.}
    \vspace{-1em}
    \label{fig:role}
\end{wrapfigure}
\textbf{\ourmethod Consistently Delivers Role-Aware Memory.} 
As shown in \Cref{fig:role}, \ourmethod consistently generates role-specific latent memories across both in-domain and out-of-domain datasets, as well as seen and unseen MAS. In the left panel (in-domain KodCode, seen MAS AutoGen), user-proxy and assistant memories form two clearly separated clusters. In the right panel (out-of-domain BigCodeBench, unseen MAS CAMEL), the role-specific memories remain well separated, demonstrating \ourmethod’s ability to avoid homogeneous memory even in entirely novel task domains, agent roles, and collaboration patterns.

\begin{wrapfigure}{t}{0.5\textwidth}
    \vspace{-1.0em}
    \centering
    \includegraphics[width=0.5\columnwidth]{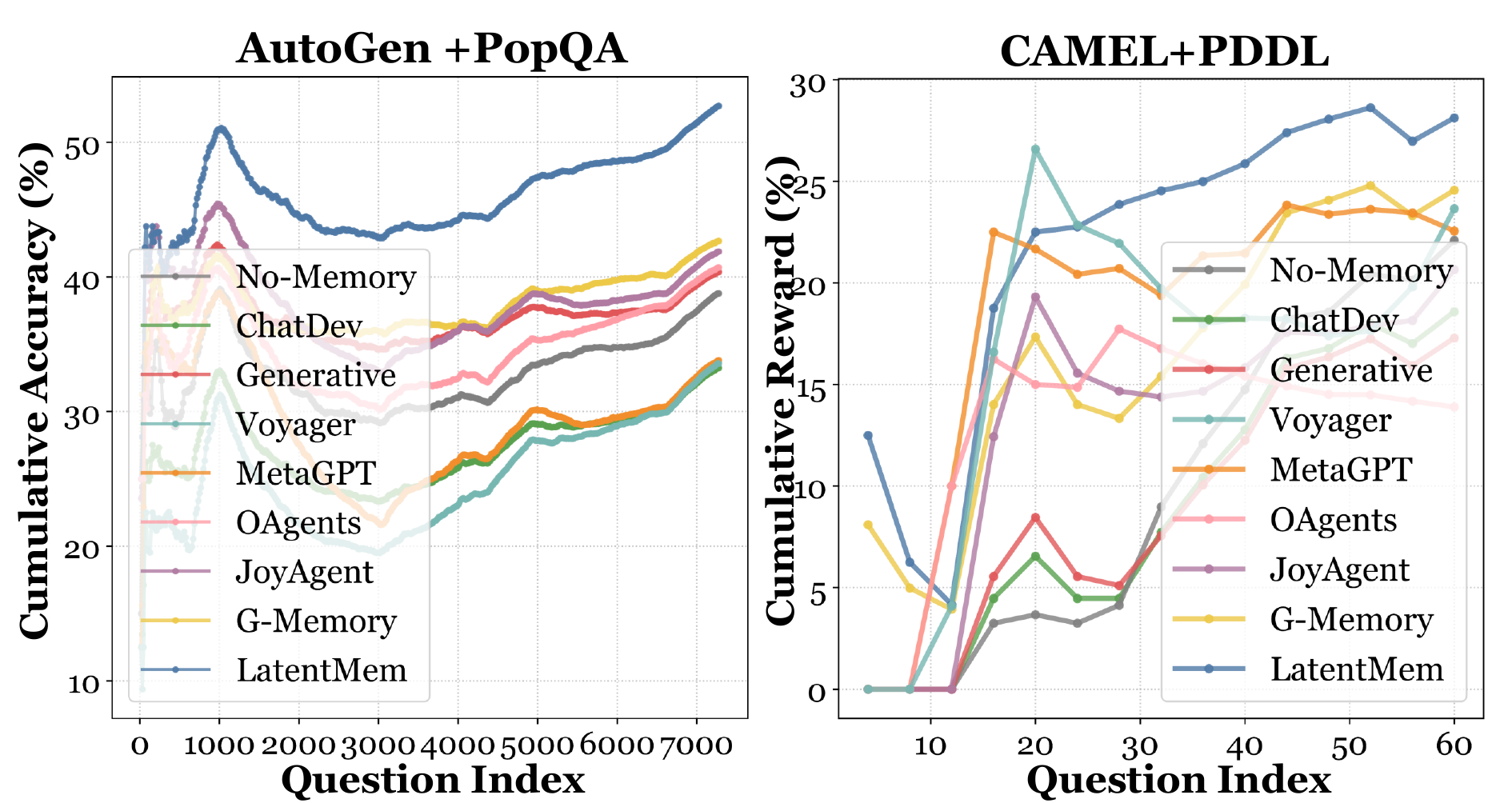}
    \caption{Evolution of cumulative accuracy (reward) across question indices. The cumulative accuracy at index $i$ is defined as the average accuracy (reward) over the first $i$ questions.}
    \vspace{-1em}
    \label{fig:evolve}
\end{wrapfigure}
\textbf{\ourmethod Scales Efficiently as Task Horizon Expands.} 
We visualize the cumulative gains of different memory systems as tasks progress, specifically by tracking their impact on cumulative accuracy. As shown in \Cref{fig:evolve}, \ourmethod steadily improves as more experiences are collected, surpassing all baselines that rely on complex, multi-granularity memory. Although early performance exhibits higher variance due to limited samples, \ourmethod quickly stabilizes and continues to improve, demonstrating its ability to efficiently distill high-utility, transferable knowledge from past interaction trajectories, which can then be leveraged to guide the reasoning process of MAS.

\subsection{Sensitivity \& Ablation Study.}\label{sec:ablation}

\begin{wrapfigure}{r}{0.5\textwidth}
    \vspace{-1em}
    \centering
    \includegraphics[width=0.5\columnwidth]{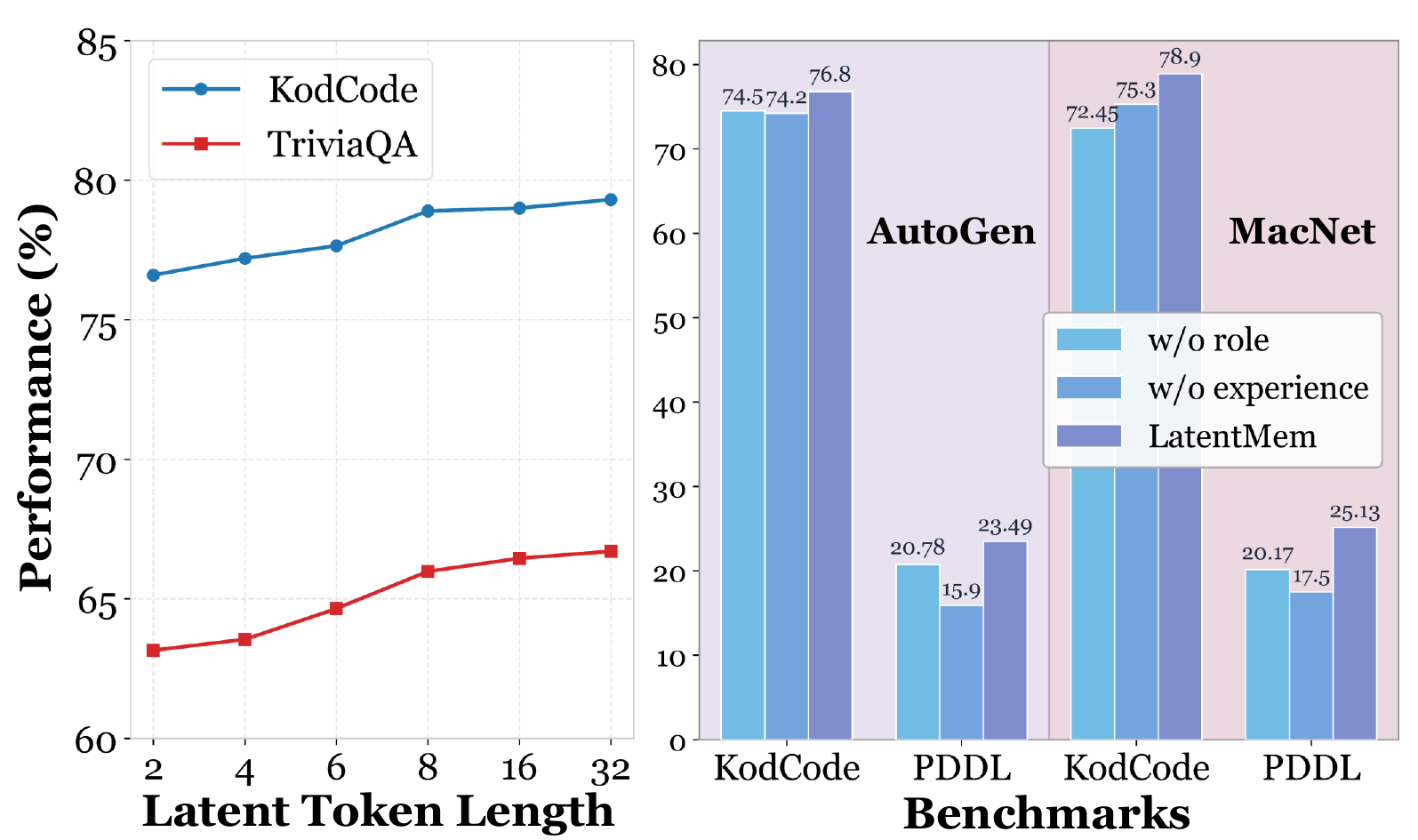}
    \caption{(\textit{Left}) Sensitivity of model performance to the latent memory length $L'$. (\textit{Right}) Ablation results highlighting the impact of the memory composer and the experience bank.}
    \vspace{-2em}
    \label{fig:sen_abl}
\end{wrapfigure}
\textbf{Sensitivity Analysis.}
We analyze the sensitivity of \ourmethod to two key hyperparameters: the latent memory length $L'$ and the number of relevant trajectories $K$. As shown in \Cref{fig:sen_abl} (\textit{Left}), performance generally improves with larger $L'$, but with diminishing returns; balancing accuracy and computational cost, we set $L' = 8$. The effect of $K$ is detailed in Appendix \ref{app:sensitivity}: while baselines such as G-Memory (see Appendix \ref{app:baseline} for more details) degrades when $K>3$ due to \textit{information overload}, \ourmethod consistently improves on both TriviaQA and KodCode, demonstrating its ability to distill useful information from redundant trajectories via latent memory.

\textbf{Component Ablation.}
We present ablation studies of \ourmethod in \Cref{fig:sen_abl} (\textit{Right}), where we introduce two variants: \textit{without role} and \textit{without experience}, corresponding to the removal of agent profile guidance ($\gamma$ in \cref{eq:composer}) and the disabling of real-time updates in the experience bank (as in \cref{eq:exp_update}), respectively. When the memory composer no longer receives agent profiles, resulting in identical latent memories across agents, performance drops slightly for simple MAS such as AutoGen (2.30\% on KodCode) and more substantially for complex MAS like MacNet (6.45\%), highlighting the importance of agent-aware memory. Disabling real-time updates in the experience bank leads to minor performance degradation on KodCode (3.60\% on MacNet) but a larger drop on PDDL (7.63\%), demonstrating its crucial role in adapting to complex task distributions. These results underscore the contributions of both components to the overall effectiveness of \ourmethod.

\begin{figure*}[!t]
    \centering
    \includegraphics[width=\linewidth]{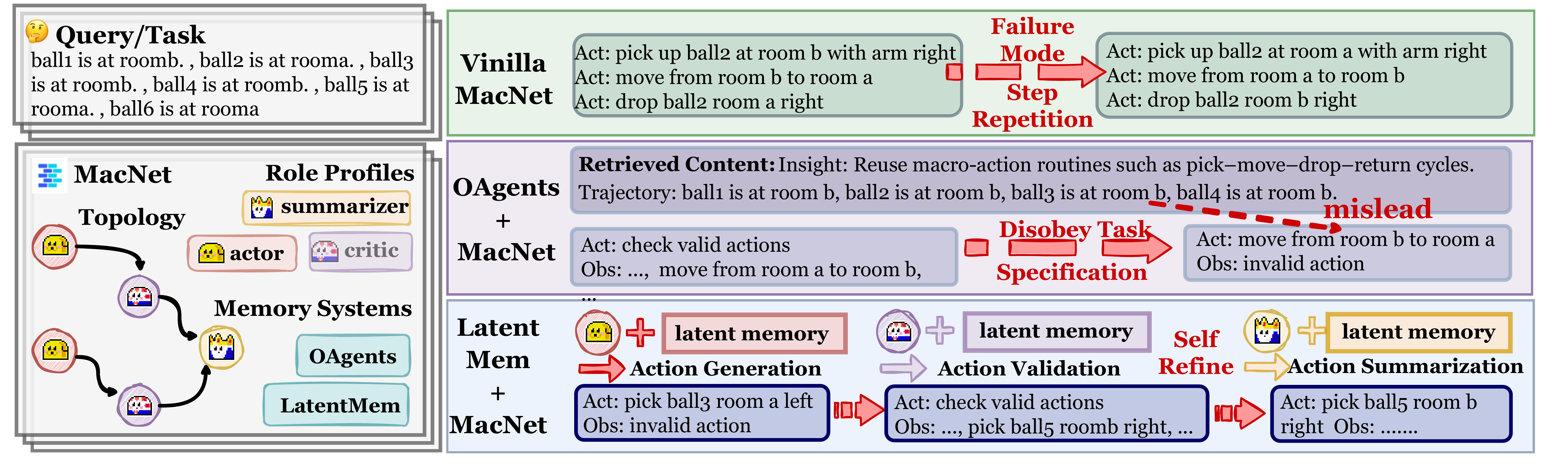}
    \caption{Case study of \ourmethod. By leveraging role-aware and compact latent memory, \ourmethod prevents common MAS issues such as step repetition and blindly following retrieved trajectories, while enabling role-aware coordination and self-correction.}
    \vspace{-1em}
    \label{fig:case_study}
\end{figure*}
\subsection{Case Study}
\Cref{fig:case_study} shows that \ourmethod, by providing role-aware memory, can prevent or promptly correct common error patterns in MAS. Vanilla MacNet often suffers from step repetition, while MacNet with OAgents blindly follow the retrieved trajectories, violating task specifications. In contrast, \ourmethod’s high-level, role-aware latent memory enables agents to reinforce role compliance and coordinate effectively, allowing the MAS to self-correct short-term errors and complete tasks successfully. Detailed trajectories and error analyses are in Appendix~\ref{app:case_study}.

\section{Conclusion}
In this work, we present \ourmethod, a latent memory framework for multi-agent systems that enables role-aware and token-efficient memory customization. By leveraging a lightweight experience bank and a learnable memory composer, each agent receives its latent memories distilled from raw trajectories, naturally reinforcing role compliance and enhancing coordination. We further introduce Latent Memory Policy Optimization, which encourages the composer to produce transferable, high-utility latent representations, enhancing generalization across diverse task domains and MAS frameworks. Extensive experiments on six benchmarks and four MAS frameworks demonstrate that \ourmethod achieves substantial performance gains, robust generalization, and high efficiency, while effectively mitigating memory homogenization and information overload.

\section*{Impact Statement}



\paragraph{Ethical Considerations.}
This study focuses on developing and evaluating multi-agent memory mechanisms using publicly accessible benchmarks and datasets. It does not involve the collection, processing, or deployment of private, personal, or sensitive user information, and all experiments are carried out in controlled, offline research environments. Consequently, we do not foresee any major ethical risks associated with this work.

\paragraph{Societal Implications.}
The methods proposed in this paper seek to enhance the robustness and reliability of LLM-based multi-agent systems, with potential benefits for applications including assistive robotics, information organization, and long-term decision-making. However, more powerful memory capabilities may also increase the potential for misuse if such systems are deployed without proper safeguards. As such, we position this work primarily as a research contribution and stress that real-world deployment should be accompanied by appropriate oversight, safety assessments, and compliance with legal and ethical standards.


\bibliography{references}

@article{ye2025x,
  title={X-MAS: Towards Building Multi-Agent Systems with Heterogeneous LLMs},
  author={Ye, Rui and Liu, Xiangrui and Wu, Qimin and Pang, Xianghe and Yin, Zhenfei and Bai, Lei and Chen, Siheng},
  journal={arXiv preprint arXiv:2505.16997},
  year={2025}
}

@article{zhang2024cut,
  title={Cut the crap: An economical communication pipeline for llm-based multi-agent systems},
  author={Zhang, Guibin and Yue, Yanwei and Li, Zhixun and Yun, Sukwon and Wan, Guancheng and Wang, Kun and Cheng, Dawei and Yu, Jeffrey Xu and Chen, Tianlong},
  journal={arXiv preprint arXiv:2410.02506},
  year={2024}
}

@misc{zou2025latentcollaborationmultiagentsystems,
      title={Latent Collaboration in Multi-Agent Systems}, 
      author={Jiaru Zou and Xiyuan Yang and Ruizhong Qiu and Gaotang Li and Katherine Tieu and Pan Lu and Ke Shen and Hanghang Tong and Yejin Choi and Jingrui He and James Zou and Mengdi Wang and Ling Yang},
      year={2025},
      eprint={2511.20639},
      archivePrefix={arXiv},
      primaryClass={cs.CL},
      url={https://arxiv.org/abs/2511.20639}, 
}

@misc{li2026seekdarkreasoningtesttime,
      title={Seek in the Dark: Reasoning via Test-Time Instance-Level Policy Gradient in Latent Space}, 
      author={Hengli Li and Chenxi Li and Tong Wu and Xuekai Zhu and Yuxuan Wang and Zhaoxin Yu and Eric Hanchen Jiang and Song-Chun Zhu and Zixia Jia and Ying Nian Wu and Zilong Zheng},
      year={2026},
      eprint={2505.13308},
      archivePrefix={arXiv},
      primaryClass={cs.LG},
      url={https://arxiv.org/abs/2505.13308}, 
}

@misc{xu2025softcotsoftchainofthoughtefficient,
      title={SoftCoT: Soft Chain-of-Thought for Efficient Reasoning with LLMs}, 
      author={Yige Xu and Xu Guo and Zhiwei Zeng and Chunyan Miao},
      year={2025},
      eprint={2502.12134},
      archivePrefix={arXiv},
      primaryClass={cs.CL},
      url={https://arxiv.org/abs/2502.12134}, 
}

@misc{zhang2025memgenweavinggenerativelatent,
      title={MemGen: Weaving Generative Latent Memory for Self-Evolving Agents}, 
      author={Guibin Zhang and Muxin Fu and Shuicheng Yan},
      year={2025},
      eprint={2509.24704},
      archivePrefix={arXiv},
      primaryClass={cs.CL},
      url={https://arxiv.org/abs/2509.24704}, 
}

@article{tang2025agent,
  title={Agent kb: Leveraging cross-domain experience for agentic problem solving},
  author={Tang, Xiangru and Qin, Tianrui and Peng, Tianhao and Zhou, Ziyang and Shao, Daniel and Du, Tingting and Wei, Xinming and Xia, Peng and Wu, Fang and Zhu, He and others},
  journal={arXiv preprint arXiv:2507.06229},
  year={2025}
}

@article{wu2025evolver,
  title={Evolver: Self-evolving llm agents through an experience-driven lifecycle},
  author={Wu, Rong and Wang, Xiaoman and Mei, Jianbiao and Cai, Pinlong and Fu, Daocheng and Yang, Cheng and Wen, Licheng and Yang, Xuemeng and Shen, Yufan and Wang, Yuxin and others},
  journal={arXiv preprint arXiv:2510.16079},
  year={2025}
}

@inproceedings{chen2023agentverse,
  title={Agentverse: Facilitating multi-agent collaboration and exploring emergent behaviors},
  author={Chen, Weize and Su, Yusheng and Zuo, Jingwei and Yang, Cheng and Yuan, Chenfei and Chan, Chi-Min and Yu, Heyang and Lu, Yaxi and Hung, Yi-Hsin and Qian, Chen and others},
  booktitle={The Twelfth International Conference on Learning Representations},
  year={2023}
}

@article{yue2025masrouter,
  title={Masrouter: Learning to route llms for multi-agent systems},
  author={Yue, Yanwei and Zhang, Guibin and Liu, Boyang and Wan, Guancheng and Wang, Kun and Cheng, Dawei and Qi, Yiyan},
  journal={arXiv preprint arXiv:2502.11133},
  year={2025}
}

@article{yang2025revisiting,
  title={Revisiting Multi-Agent Debate as Test-Time Scaling: A Systematic Study of Conditional Effectiveness},
  author={Yang, Yongjin and Yi, Euiin and Ko, Jongwoo and Lee, Kimin and Jin, Zhijing and Yun, Se-Young},
  journal={arXiv preprint arXiv:2505.22960},
  year={2025}
}

@article{zhang2024can,
  title={Can LLMs beat humans in debating? A dynamic multi-agent framework for competitive debate},
  author={Zhang, Yiqun and Yang, Xiaocui and Feng, Shi and Wang, Daling and Zhang, Yifei and Song, Kaisong},
  journal={arXiv preprint arXiv:2408.04472},
  year={2024}
}

@article{wang2025talk,
  title={Talk structurally, act hierarchically: A collaborative framework for llm multi-agent systems},
  author={Wang, Zhao and Moriyama, Sota and Wang, Wei-Yao and Gangopadhyay, Briti and Takamatsu, Shingo},
  journal={arXiv preprint arXiv:2502.11098},
  year={2025}
}

@article{zhang2025g,
  title={G-Memory: Tracing Hierarchical Memory for Multi-Agent Systems},
  author={Zhang, Guibin and Fu, Muxin and Wan, Guancheng and Yu, Miao and Wang, Kun and Yan, Shuicheng},
  journal={arXiv preprint arXiv:2506.07398},
  year={2025}
}

@article{cemri2025multi,
  title={Why do multi-agent llm systems fail?},
  author={Cemri, Mert and Pan, Melissa Z and Yang, Shuyi and Agrawal, Lakshya A and Chopra, Bhavya and Tiwari, Rishabh and Keutzer, Kurt and Parameswaran, Aditya and Klein, Dan and Ramchandran, Kannan and others},
  journal={arXiv preprint arXiv:2503.13657},
  year={2025}
}

@article{wang2025mirix,
  title={Mirix: Multi-agent memory system for llm-based agents},
  author={Wang, Yu and Chen, Xi},
  journal={arXiv preprint arXiv:2507.07957},
  year={2025}
}

@article{qian2024scaling,
  title={Scaling large language model-based multi-agent collaboration},
  author={Qian, Chen and Xie, Zihao and Wang, Yifei and Liu, Wei and Zhu, Kunlun and Xia, Hanchen and Dang, Yufan and Du, Zhuoyun and Chen, Weize and Yang, Cheng and others},
  journal={arXiv preprint arXiv:2406.07155},
  year={2024}
}

@inproceedings{zhao2024expel,
  title={Expel: Llm agents are experiential learners},
  author={Zhao, Andrew and Huang, Daniel and Xu, Quentin and Lin, Matthieu and Liu, Yong-Jin and Huang, Gao},
  booktitle={Proceedings of the AAAI Conference on Artificial Intelligence},
  volume={38},
  pages={19632--19642},
  year={2024}
}

@article{liu2025joyagent,
  title={JoyAgent-JDGenie: Technical Report on the GAIA},
  author={Liu, Jiarun and Xu, Shiyue and Liu, Shangkun and Li, Yang and Liu, Wen and Liu, Min and Zhou, Xiaoqing and Wang, Hanmin and Jia, Shilin and Tian, Shaohua and others},
  journal={arXiv preprint arXiv:2510.00510},
  year={2025}
}

@article{zhu2025oagents,
  title={Oagents: An empirical study of building effective agents},
  author={Zhu, He and Qin, Tianrui and Zhu, King and Huang, Heyuan and Guan, Yeyi and Xia, Jinxiang and Yao, Yi and Li, Hanhao and Wang, Ningning and Liu, Pai and others},
  journal={arXiv preprint arXiv:2506.15741},
  year={2025}
}

@article{zhang2025memevolve,
  title={Memevolve: Meta-evolution of agent memory systems},
  author={Zhang, Guibin and Ren, Haotian and Zhan, Chong and Zhou, Zhenhong and Wang, Junhao and Zhu, He and Zhou, Wangchunshu and Yan, Shuicheng},
  journal={arXiv preprint arXiv:2512.18746},
  year={2025}
}

@article{han2025legomem,
  title={Legomem: Modular procedural memory for multi-agent llm systems for workflow automation},
  author={Han, Dongge and Couturier, Camille and Diaz, Daniel Madrigal and Zhang, Xuchao and R{\"u}hle, Victor and Rajmohan, Saravan},
  journal={arXiv preprint arXiv:2510.04851},
  year={2025}
}

@article{wu2025human,
  title={From human memory to ai memory: A survey on memory mechanisms in the era of llms},
  author={Wu, Yaxiong and Liang, Sheng and Zhang, Chen and Wang, Yichao and Zhang, Yongyue and Guo, Huifeng and Tang, Ruiming and Liu, Yong},
  journal={arXiv preprint arXiv:2504.15965},
  year={2025}
}

@article{hu2025memory,
  title={Memory in the Age of AI Agents},
  author={Hu, Yuyang and Liu, Shichun and Yue, Yanwei and Zhang, Guibin and Liu, Boyang and Zhu, Fangyi and Lin, Jiahang and Guo, Honglin and Dou, Shihan and Xi, Zhiheng and others},
  journal={arXiv preprint arXiv:2512.13564},
  year={2025}
}

@article{subramaniam2025multiagent,
  title={Multiagent finetuning: Self improvement with diverse reasoning chains},
  author={Subramaniam, Vighnesh and Du, Yilun and Tenenbaum, Joshua B and Torralba, Antonio and Li, Shuang and Mordatch, Igor},
  journal={arXiv preprint arXiv:2501.05707},
  year={2025}
}

@article{wang2020microsoft,
  title={Microsoft academic graph: When experts are not enough},
  author={Wang, Kuansan and Shen, Zhihong and Huang, Chiyuan and Wu, Chieh-Han and Dong, Yuxiao and Kanakia, Anshul},
  journal={Quantitative Science Studies},
  volume={1},
  number={1},
  pages={396--413},
  year={2020},
  publisher={MIT Press One Rogers Street, Cambridge, MA 02142-1209, USA journals-info~…}
}

@article{sutton2019bitter,
  title={The bitter lesson},
  author={Sutton, Richard},
  journal={Incomplete Ideas (blog)},
  volume={13},
  number={1},
  pages={38},
  year={2019}
}

@article{yu2025dapo,
  title={Dapo: An open-source llm reinforcement learning system at scale},
  author={Yu, Qiying and Zhang, Zheng and Zhu, Ruofei and Yuan, Yufeng and Zuo, Xiaochen and Yue, Yu and Dai, Weinan and Fan, Tiantian and Liu, Gaohong and Liu, Lingjun and others},
  journal={arXiv preprint arXiv:2503.14476},
  year={2025}
}

@article{zheng2026lifelong,
  title={Lifelong learning of large language model based agents: A roadmap},
  author={Zheng, Junhao and Shi, Chengming and Cai, Xidi and Li, Qiuke and Zhang, Duzhen and Li, Chenxing and Yu, Dong and Ma, Qianli},
  journal={IEEE Transactions on Pattern Analysis and Machine Intelligence},
  year={2026},
  publisher={IEEE}
}

@article{tomilin2025meal,
  title={MEAL: A Benchmark for Continual Multi-Agent Reinforcement Learning},
  author={Tomilin, Tristan and Boogaard, Luka van den and Garcin, Samuel and Grooten, Bram and Fang, Meng and Du, Yali and Pechenizkiy, Mykola},
  journal={arXiv preprint arXiv:2506.14990},
  year={2025}
}

@article{wang2020minilm,
  title={Minilm: Deep self-attention distillation for task-agnostic compression of pre-trained transformers},
  author={Wang, Wenhui and Wei, Furu and Dong, Li and Bao, Hangbo and Yang, Nan and Zhou, Ming},
  journal={Advances in Neural Information Processing Systems},
  volume={33},
  pages={5776--5788},
  year={2020}
}

@inproceedings{joshi-etal-2017-triviaqa,
    title = "{T}rivia{QA}: A Large Scale Distantly Supervised Challenge Dataset for Reading Comprehension",
    author = "Joshi, Mandar  and
      Choi, Eunsol  and
      Weld, Daniel  and
      Zettlemoyer, Luke",
    editor = "Barzilay, Regina  and
      Kan, Min-Yen",
    booktitle = "Proceedings of the 55th Annual Meeting of the Association for Computational Linguistics (Volume 1: Long Papers)",
    month = jul,
    year = "2017",
    address = "Vancouver, Canada",
    publisher = "Association for Computational Linguistics",
    url = "https://aclanthology.org/P17-1147/",
    doi = "10.18653/v1/P17-1147",
    pages = "1601--1611"
}

@inproceedings{mallen2023not,
  title={When not to trust language models: Investigating effectiveness of parametric and non-parametric memories},
  author={Mallen, Alex and Asai, Akari and Zhong, Victor and Das, Rajarshi and Khashabi, Daniel and Hajishirzi, Hannaneh},
  booktitle={Proceedings of the 61st Annual Meeting of the Association for Computational Linguistics (Volume 1: Long Papers)},
  pages={9802--9822},
  year={2023}
}

@article{xu2025kodcode,
  title={Kodcode: A diverse, challenging, and verifiable synthetic dataset for coding},
  author={Xu, Zhangchen and Liu, Yang and Yin, Yueqin and Zhou, Mingyuan and Poovendran, Radha},
  journal={arXiv preprint arXiv:2503.02951},
  year={2025}
}

@article{jain2024livecodebench,
  title={Livecodebench: Holistic and contamination free evaluation of large language models for code},
  author={Jain, Naman and Han, King and Gu, Alex and Li, Wen-Ding and Yan, Fanjia and Zhang, Tianjun and Wang, Sida and Solar-Lezama, Armando and Sen, Koushik and Stoica, Ion},
  journal={arXiv preprint arXiv:2403.07974},
  year={2024}
}

@article{geva2021did,
  title={Did aristotle use a laptop? a question answering benchmark with implicit reasoning strategies},
  author={Geva, Mor and Khashabi, Daniel and Segal, Elad and Khot, Tushar and Roth, Dan and Berant, Jonathan},
  journal={Transactions of the Association for Computational Linguistics},
  volume={9},
  pages={346--361},
  year={2021},
  publisher={MIT Press One Rogers Street, Cambridge, MA 02142-1209, USA journals-info~…}
}

@inproceedings{silver2024generalized,
  title={Generalized planning in pddl domains with pretrained large language models},
  author={Silver, Tom and Dan, Soham and Srinivas, Kavitha and Tenenbaum, Joshua B and Kaelbling, Leslie and Katz, Michael},
  booktitle={Proceedings of the AAAI conference on artificial intelligence},
  volume={38},
  pages={20256--20264},
  year={2024}
}

@article{wang2023voyager,
  title={Voyager: An open-ended embodied agent with large language models},
  author={Wang, Guanzhi and Xie, Yuqi and Jiang, Yunfan and Mandlekar, Ajay and Xiao, Chaowei and Zhu, Yuke and Fan, Linxi and Anandkumar, Anima},
  journal={arXiv preprint arXiv:2305.16291},
  year={2023}
}

@inproceedings{park2023generative,
  title={Generative agents: Interactive simulacra of human behavior},
  author={Park, Joon Sung and O'Brien, Joseph and Cai, Carrie Jun and Morris, Meredith Ringel and Liang, Percy and Bernstein, Michael S},
  booktitle={Proceedings of the 36th annual acm symposium on user interface software and technology},
  pages={1--22},
  year={2023}
}

@inproceedings{hong2023metagpt,
  title={MetaGPT: Meta programming for a multi-agent collaborative framework},
  author={Hong, Sirui and Zhuge, Mingchen and Chen, Jonathan and Zheng, Xiawu and Cheng, Yuheng and Wang, Jinlin and Zhang, Ceyao and Wang, Zili and Yau, Steven Ka Shing and Lin, Zijuan and others},
  booktitle={The twelfth international conference on learning representations},
  year={2023}
}

@inproceedings{qian2024chatdev,
  title={Chatdev: Communicative agents for software development},
  author={Qian, Chen and Liu, Wei and Liu, Hongzhang and Chen, Nuo and Dang, Yufan and Li, Jiahao and Yang, Cheng and Chen, Weize and Su, Yusheng and Cong, Xin and others},
  booktitle={Proceedings of the 62nd Annual Meeting of the Association for Computational Linguistics (Volume 1: Long Papers)},
  pages={15174--15186},
  year={2024}
}

@inproceedings{wu2024autogen,
  title={Autogen: Enabling next-gen LLM applications via multi-agent conversations},
  author={Wu, Qingyun and Bansal, Gagan and Zhang, Jieyu and Wu, Yiran and Li, Beibin and Zhu, Erkang and Jiang, Li and Zhang, Xiaoyun and Zhang, Shaokun and Liu, Jiale and others},
  booktitle={First Conference on Language Modeling},
  year={2024}
}

@article{li2023camel,
  title={Camel: Communicative agents for" mind" exploration of large language model society},
  author={Li, Guohao and Hammoud, Hasan and Itani, Hani and Khizbullin, Dmitrii and Ghanem, Bernard},
  journal={Advances in Neural Information Processing Systems},
  volume={36},
  pages={51991--52008},
  year={2023}
}

@inproceedings{liu2024dynamic,
  title={A dynamic llm-powered agent network for task-oriented agent collaboration},
  author={Liu, Zijun and Zhang, Yanzhe and Li, Peng and Liu, Yang and Yang, Diyi},
  booktitle={First Conference on Language Modeling},
  year={2024}
}

@misc{marti2025,
  title={MARTI: A Framework for Multi-Agent LLM Systems Reinforced Training and Inference},
  author={Kaiyan Zhang and Runze Liu and Xuekai Zhu and Kai Tian and Sihang Zeng and Guoli Jia and Yuchen Fan and Xingtai Lv and Yuxin Zuo and Che Jiang and Ziyang Liu and Jianyu Wang and Yuru Wang and Ruotong Zhao and Ermo Hua and Yibo Wang and Shijie Wang and Junqi Gao and Xinwei Long and Youbang Sun and Zhiyuan Ma and Ganqu Cui and Lei Bai and Ning Ding and Biqing Qi and Bowen Zhou},
  year={2025},
  institution={Tsinghua University and Shanghai AI Lab},
  url={https://github.com/TsinghuaC3I/MARTI}
}

@article{zhang2024chain,
  title={Chain of agents: Large language models collaborating on long-context tasks},
  author={Zhang, Yusen and Sun, Ruoxi and Chen, Yanfei and Pfister, Tomas and Zhang, Rui and Arik, Sercan},
  journal={Advances in Neural Information Processing Systems},
  volume={37},
  pages={132208--132237},
  year={2024}
}

@article{zhou2025multi,
  title={Multi-agent design: Optimizing agents with better prompts and topologies},
  author={Zhou, Han and Wan, Xingchen and Sun, Ruoxi and Palangi, Hamid and Iqbal, Shariq and Vuli{\'c}, Ivan and Korhonen, Anna and Ar{\i}k, Sercan {\"O}},
  journal={arXiv preprint arXiv:2502.02533},
  year={2025}
}

@article{zhang2025multi,
  title={Multi-agent architecture search via agentic supernet},
  author={Zhang, Guibin and Niu, Luyang and Fang, Junfeng and Wang, Kun and Bai, Lei and Wang, Xiang},
  journal={arXiv preprint arXiv:2502.04180},
  year={2025}
}

@article{yang2025agentnet,
  title={Agentnet: Decentralized evolutionary coordination for llm-based multi-agent systems},
  author={Yang, Yingxuan and Chai, Huacan and Shao, Shuai and Song, Yuanyi and Qi, Siyuan and Rui, Renting and Zhang, Weinan},
  journal={arXiv preprint arXiv:2504.00587},
  year={2025}
}

@article{zhang2024g,
  title={G-designer: Architecting multi-agent communication topologies via graph neural networks},
  author={Zhang, Guibin and Yue, Yanwei and Sun, Xiangguo and Wan, Guancheng and Yu, Miao and Fang, Junfeng and Wang, Kun and Chen, Tianlong and Cheng, Dawei},
  journal={arXiv preprint arXiv:2410.11782},
  year={2024}
}

@article{chang2024agentboard,
  title={Agentboard: An analytical evaluation board of multi-turn llm agents},
  author={Chang, Ma and Zhang, Junlei and Zhu, Zhihao and Yang, Cheng and Yang, Yujiu and Jin, Yaohui and Lan, Zhenzhong and Kong, Lingpeng and He, Junxian},
  journal={Advances in neural information processing systems},
  volume={37},
  pages={74325--74362},
  year={2024}
}

@article{maaten2008visualizing,
  title={Visualizing data using t-SNE},
  author={Maaten, Laurens van der and Hinton, Geoffrey},
  journal={Journal of machine learning research},
  volume={9},
  number={Nov},
  pages={2579--2605},
  year={2008}
}

@article{shao2024deepseekmath,
  title={Deepseekmath: Pushing the limits of mathematical reasoning in open language models},
  author={Shao, Zhihong and Wang, Peiyi and Zhu, Qihao and Xu, Runxin and Song, Junxiao and Bi, Xiao and Zhang, Haowei and Zhang, Mingchuan and Li, YK and Wu, Yang and others},
  journal={arXiv preprint arXiv:2402.03300},
  year={2024}
}

@article{qu2025survey,
  title={A survey of efficient reasoning for large reasoning models: Language, multimodality, and beyond},
  author={Qu, Xiaoye and Li, Yafu and Su, Zhaochen and Sun, Weigao and Yan, Jianhao and Liu, Dongrui and Cui, Ganqu and Liu, Daizong and Liang, Shuxian and He, Junxian and others},
  journal={arXiv preprint arXiv:2503.21614},
  year={2025}
}

@article{yang2026toward,
  title={Toward Efficient Agents: Memory, Tool learning, and Planning},
  author={Yang, Xiaofang and Li, Lijun and Zhou, Heng and Zhu, Tong and Qu, Xiaoye and Fan, Yuchen and Wei, Qianshan and Ye, Rui and Kang, Li and Qin, Yiran and others},
  journal={arXiv preprint arXiv:2601.14192},
  year={2026}
}

@article{zhang2025survey,
  title={A survey of reinforcement learning for large reasoning models},
  author={Zhang, Kaiyan and Zuo, Yuxin and He, Bingxiang and Sun, Youbang and Liu, Runze and Jiang, Che and Fan, Yuchen and Tian, Kai and Jia, Guoli and Li, Pengfei and others},
  journal={arXiv preprint arXiv:2509.08827},
  year={2025}
}

@article{guo2024large,
  title={Large language model based multi-agents: A survey of progress and challenges},
  author={Guo, Taicheng and Chen, Xiuying and Wang, Yaqi and Chang, Ruidi and Pei, Shichao and Chawla, Nitesh V and Wiest, Olaf and Zhang, Xiangliang},
  journal={arXiv preprint arXiv:2402.01680},
  year={2024}
}

@article{li2024survey,
  title={A survey on LLM-based multi-agent systems: workflow, infrastructure, and challenges},
  author={Li, Xinyi and Wang, Sai and Zeng, Siqi and Wu, Yu and Yang, Yi},
  journal={Vicinagearth},
  year={2024}
}

@article{tran2025multi,
  title={Multi-agent collaboration mechanisms: A survey of llms},
  author={Tran, Khanh-Tung and Dao, Dung and Nguyen, Minh-Duong and Pham, Quoc-Viet and O'Sullivan, Barry and Nguyen, Hoang D},
  journal={arXiv preprint arXiv:2501.06322},
  year={2025}
}

@article{du2023improving,
  title={Improving factuality and reasoning in language models through multiagent debate, 2023},
  author={Du, Yilun and Li, Shuang and Torralba, Antonio and Tenenbaum, Joshua B and Mordatch, Igor},
  journal={arXiv preprint arXiv:2305.14325},
  year={2023}
}

@article{liang2023encouraging,
  title={Encouraging divergent thinking in large language models through multi-agent debate},
  author={Liang, Tian and He, Zhiwei and Jiao, Wenxiang and Wang, Xing and Wang, Yan and Wang, Rui and Yang, Yujiu and Shi, Shuming and Tu, Zhaopeng},
  journal={arXiv preprint arXiv:2305.19118},
  year={2023}
}

@article{wang2024moa,
  title={Mixture-of-agents enhances large language model capabilities},
  author={Wang, Junlin and Wang, Jue and Athiwaratkun, Ben and others},
  journal={arXiv preprint arXiv:2406.04692},
  year={2024}
}

@inproceedings{zhuge2024gptswarm,
  title={Gptswarm: Language agents as optimizable graphs},
  author={Zhuge, Mingchen and Wang, Wenyi and Kirsch, Louis and Faccio, Francesco and Khizbullin, Dmitrii and Schmidhuber, J{\"u}rgen},
  booktitle={Forty-first International Conference on Machine Learning},
  year={2024}
}

@article{zhou2025reso,
  title={Reso: A reward-driven self-organizing llm-based multi-agent system for reasoning tasks},
  author={Zhou, Heng and Geng, Hejia and Xue, Xiangyuan and Kang, Li and Qin, Yiran and Wang, Zhiyong and Yin, Zhenfei and Bai, Lei},
  journal={arXiv preprint arXiv:2503.02390},
  year={2025}
}

@article{rezazadeh2025collaborative,
  title={Collaborative Memory: Multi-User Memory Sharing in LLM Agents with Dynamic Access Control},
  author={Rezazadeh, Alireza and Li, Zichao and Lou, Ange and Zhao, Yuying and Wei, Wei and Bao, Yujia},
  journal={arXiv preprint arXiv:2505.18279},
  year={2025}
}

@inproceedings{yin2023exchange,
  title={Exchange-of-thought: Enhancing large language model capabilities through cross-model communication},
  author={Yin, Zhangyue and Sun, Qiushi and Chang, Cheng and Guo, Qipeng and Dai, Junqi and Huang, Xuan-Jing and Qiu, Xipeng},
  booktitle={Proceedings of the 2023 Conference on Empirical Methods in Natural Language Processing},
  pages={15135--15153},
  year={2023}
}

@article{qian2023communicative,
  title={Communicative agents for software development},
  author={Qian, Chen and Cong, Xin and Yang, Cheng and Chen, Weize and Su, Yusheng and Xu, Juyuan and Liu, Zhiyuan and Sun, Maosong},
  journal={arXiv preprint arXiv:2307.07924},
  volume={6},
  number={3},
  pages={1},
  year={2023}
}

@article{xu2025mem,
  title={A-mem: Agentic memory for llm agents},
  author={Xu, Wujiang and Liang, Zujie and Mei, Kai and Gao, Hang and Tan, Juntao and Zhang, Yongfeng},
  journal={arXiv preprint arXiv:2502.12110},
  year={2025}
}

@article{hu2022lora,
  title={Lora: Low-rank adaptation of large language models.},
  author={Hu, Edward J and Shen, Yelong and Wallis, Phillip and Allen-Zhu, Zeyuan and Li, Yuanzhi and Wang, Shean and Wang, Lu and Chen, Weizhu and others},
  journal={ICLR},
  volume={1},
  pages={3},
  year={2022}
}

@misc{fu2025cachetocachedirectsemanticcommunication,
      title={Cache-to-Cache: Direct Semantic Communication Between Large Language Models}, 
      author={Tianyu Fu and Zihan Min and Hanling Zhang and Jichao Yan and Guohao Dai and Wanli Ouyang and Yu Wang},
      year={2025},
      eprint={2510.03215},
      archivePrefix={arXiv},
      primaryClass={cs.CL},
      url={https://arxiv.org/abs/2510.03215}, 
}

\newpage
\appendix
\onecolumn



\section{More Related Work}
This work is closely related to prior studies on latent reasoning, including but not limited to MemGen~\citep{zhang2025memgenweavinggenerativelatent}, SoftCoT~\citep{xu2025softcotsoftchainofthoughtefficient}, and LatentSeek~\citep{li2026seekdarkreasoningtesttime}. While these works demonstrate that latent-encoded memory can be an efficient carrier of reasoning and experience, their applicability to multi-agent memory systems remains largely unexplored. Our study addresses this gap by extending latent reasoning into the multi-agent setting and opening a new direction for structured, collaborative memory evolution. LatentMAS~\citep{zou2025latentcollaborationmultiagentsystems} and Cache-to-Cache~\citep{fu2025cachetocachedirectsemanticcommunication} are also relevant works, though they concentrate on latent communication.

\section{Experimental Details}

\subsection{Dataset Descriptions}\label{app:dataset}
In this section, we describe the datasets used in our experiments:
\begin{itemize}
    \item \textbf{TriviaQA}~\citep{joshi-etal-2017-triviaqa} is a large-scale question answering dataset comprising challenging, naturally occurring trivia questions paired with evidence documents, designed to evaluate reading comprehension and information retrieval over long contexts.
    \item \textbf{KodCode}~\cite{xu2025kodcode} is a fully synthetic coding dataset with diverse tasks, verifiable solutions, and tests, supporting both supervised fine-tuning and reinforcement learning.  
    \item \textbf{StrategyQA}~\citep{geva2021did} is a QA benchmark of yes/no questions requiring multi-step reasoning, with annotated reasoning steps and supporting evidence to evaluate implicit reasoning.  
    \item \textbf{PopQA}~\citep{mallen2023not} is a large-scale open-domain question answering dataset of entity-centric QA pairs generated from Wikidata knowledge tuples with annotated subject, relation, object, and Wikipedia page view metadata, designed to support factual QA evaluation.
    \item \textbf{BigCodeBench}~\citep{jain2024livecodebench} is a programming benchmark dataset of diverse Python coding tasks with structured prompts, canonical solutions, and unit tests, designed to evaluate and compare code generation capabilities of large language models.
    \item \textbf{PDDL}~\cite{silver2024generalized} is a game dataset from AgentBoard~\citep{chang2024agentboard}, comprising a variety of strategic games where agents use PDDL expressions to complete complex tasks.
\end{itemize}

\subsection{Baseline Setup}\label{app:baseline}
In this section, we provide detailed descriptions of each baseline used in our comparison:
\begin{itemize}

    \item \textbf{Voyager}: The Voyager memory originates from the Voyager agent~\citep{wang2023voyager}, where an embodied agent continuously explores the Minecraft environment and generates new artifacts, with memory acting as the central driver of its development. Since Voyager’s memory is originally designed for a single-agent setup, we extend it to the multi-agent setting by retrieving agent-specific histories based on each agent’s visible dialogue context. Other single-agent memory architectures are adapted similarly.
    
    \item \textbf{Generative}: This memory baseline follows Generative agent~\citep{park2023generative} and incorporates both raw observational memory and high-level reflective memory. The reflective component encodes abstract insights produced by the agent via reflection, offering a more organized and conceptual representation of its experiences.
    
    \item \textbf{JoyAgent}: Adapted from JoyAgent~\citep{liu2025joyagent}, this hierarchical memory comprises working, semantic, and procedural layers. Working memory maintains the execution context, semantic memory distills task trajectories and lessons for retrieval, and procedural memory encodes dynamic system prompts for strategic control. This structure ensures long-horizon continuity and adaptive behavior in single-agent task execution.
    
    \item \textbf{MetaGPT}: This memory design is derived from MetaGPT~\citep{hong2023metagpt}, incorporating a shared message pool that enables agents to publish and subscribe to structured information throughout the software development lifecycle. This system facilitates inside-trial coordination by allowing agents to retrieve historical context for self-correction, while also supporting cross-trial improvement through a long-term memory layer that stores feedback to refine agent constraints over time.
    
    \item \textbf{ChatDev}: This memory design is adapted from ChatDev~\citep{qian2024chatdev}, focusing solely on inside-trial memory—information stored internally during the resolution of a single task by multiple agents.

    \item \textbf{OAgents}: Based on OAgents framework~\citep{zhu2025oagents}, this design uses a hierarchical memory system: short-term for real-time tasks and long-term for fused historical insights. It employs memory summarization and vectorized retrieval to manage information density and extract relevant context. This multi-layered approach ensures long-horizon reasoning while minimizing redundancy during complex tasks.

    \item \textbf{G-Memory}: G-Memory~\citep{zhang2025g} operates through a structured lifecycle of retrieval, traversal, and updating to manage the extensive interaction histories of multi-agent systems. When a new query arrives, the system first performs a coarse-grained similarity search on the Query Graph to identify relevant historical tasks, followed by a bi-directional traversal: an upward traversal to the Insight Graph to extract high-level strategic guidance and a downward traversal to the Interaction Graph to retrieve condensed, fine-grained collaboration trajectories. These filtered memories are then customized for each agent based on their specific roles to provide actionable context without overwhelming the context window. Once the task is executed, G-Memory performs an agentic update where new collaborative logs are stored, successful or failed experiences are distilled into new nodes in the Insight Graph, and semantic edges are reinforced across the three-tier hierarchy to facilitate continuous self-evolution.
\end{itemize}

\subsection{Multi-agent System Setup}\label{app:mas}
In this section, we detail the setups of our four adopted MAS frameworks, AutoGen, MacNet, CAMEL and DyLAN.
\subsubsection{AutoGen}
AutoGen~\citep{wu2024autogen} is a widely adopted framework for orchestrating multi-agent collaboration. In this work, we adopt its A2 setting, which consists of a Retrieval-augmented User Proxy agent and a Retrieval-augmented Assistant agent, both extended from AutoGen’s built-in agents. The User Proxy agent incorporates a vector database retriever to provide relevant external context, while the Assistant agent utilizes the retrieved information to generate accurate responses. This design enables interactive retrieval and supports effective knowledge grounding for both question answering and code generation tasks.

\subsubsection{MacNet}
MacNet~\citep{qian2024scaling} is a representative work that explores decentralized and scalable multi-agent systems. Its key feature lies in the absence of a central agent; instead, it introduces edge agents, which are invoked between agent interactions to provide actionable instructions to the next agent based on the previous agent’s outputs. In our implementation, we adopt the random graph topology from MacNet, shown to be robust across diverse scenarios, and employ a total of five agents, including the edge agents.

\subsubsection{CAMEL}
CAMEL~\citep{li2023camel} is a role-playing–based communication agent framework designed to guide agents to autonomously collaborate on complex tasks with minimal human intervention via inception prompting. The framework introduces a task specifier agent that concretizes vague ideas and achieves task objectives through multi-turn, instruction-following dialogues. In our implementation, we employ four agents: three participating in the debate and one dedicated to summarization.

\subsubsection{DyLAN}
DyLAN~\citep{liu2024dynamic} is a debate-style framework similar to LLM-Debate, but incorporates a more efficient agent-wise early stopping mechanism during multi-turn interactions. DyLAN utilizes an agent selection algorithm based on an unsupervised metric, namely the Agent Importance Score, which identifies the most contributive agents through a preliminary trial tailored to the specific task. In our implementation of DyLAN, four agents engage in the debate, while an additional ranker agent evaluates their relative importance.

\subsection{Training Details}\label{app:train_details}

\subsubsection{Evaluation of Generalization}
To evaluate the generalization capabilities of different methods, we consider two dimensions: (1) whether the method can generalize to domains unseen during training, and (2) whether it can generalize to multi-agent systems unseen during training. For the in-domain and out-of-domain evaluation, we use TriviaQA, KodCode, StrategyQA, and PopQA as in-domain datasets, and BigCodeBench and PDDL as out-of-domain datasets. For MAS generalization, AutoGen and MacNet are considered as MAS encountered during training, while CAMEL and DyLAN are treated as MAS not seen during training.

\subsubsection{Training Trajectories Collection and Usage}
For all baselines and \ourmethod, we first collect data on all in-domain datasets using the training splits and in-distribution MAS. Specifically, we gather training data by running AutoGen and MacNet on TriviaQA, KodCode, StrategyQA, and PopQA. In total, we obtain 40,580 trajectories. For non-parametric baselines, such as MetaGPT~\citep{hong2023metagpt} and G-Memory~\citep{zhang2025g}, the training data is processed and stored directly in their memory modules. For parametric baselines, such as MARTI~\citep{marti2025}, the training data is used to optimize model parameters. For \ourmethod, we first follow the same procedure as non-parametric baselines: the MAS performs standard reasoning on the training data, and the resulting trajectories are stored in the experience bank (without using the memory composer at this stage). We then train the memory composer via LMPO on the same training data, enabling it to effectively transform retrieved trajectories from the experience bank into latent memories that can assist MAS reasoning.

\subsubsection{Parameter Configurations}
Here, we provide full details of the parameter settings used to implement \ourmethod. 

In practice, the memory composer is instantiated as a pretrained LLM (e.g., \llmname{Qwen3-4B-Instruct-2507} or \llmname{Llama-3.1-8B-Instruct}). Prior to training, we collect diverse datasets and raw MAS trajectories on AutoGen and MacNet using the official training splits of in-domain datasets (TriviaQA, KodCode, StrategyQA, PopQA). Training then proceeds following the same protocol applied to all baseline methods. Detailed hyperparameter configurations are reported in \Cref{tab:hyperparameters}.

\begin{table}[h]
\centering
\caption{The detailed parameter settings when implementing \ourmethod.}
\label{tab:hyperparameters}
\begin{tabular}{l c}
\hline
\textbf{Parameter} & \textbf{Setting} \\
\hline
Token limit for prompts & 10240 \\
Token limit for responses & 4096 \\
Training temperature & 1.0 \\
Evaluation temperature & 0.0 \\
Discount factor & 1.0 \\
Clipping epsilon & 0.2 \\
Weight of KL penalty & 0.0 \\
Number of training epochs & 1 \\
Number of prompt reuse & 1 \\
Macro training batch size & 32 \\
Micro training batch size & 8 \\
Macro rollout batch size & 32 \\
Micro rollout batch size & 8 \\
Optimizer name & AdamW \\
Learning rate & 1e-5 \\
Warmup ratio & 0.1 \\
Lora r & 16 \\
Lora alpha & 32 \\
Lora target modules & [q\_proj, v\_proj] \\
Lora dropout & 0.1 \\
Lora bias & None \\
Lora task type & CAUSAL\_LM \\
Gradient norm & 1.0 \\
Gradient clipping & True \\
Gradient checkpoint & True \\
Flash Attention & True \\
Mixed precision & True \\
Enable vLLM & True \\
Enable DeepSpeed & True \\
\hline
\end{tabular}
\end{table}

\section{Extra Results}\label{app:result}

\subsection{Cost Analysis}\label{app:cost}
\begin{figure}[!t]
    \centering
    \includegraphics[width=0.6\linewidth]{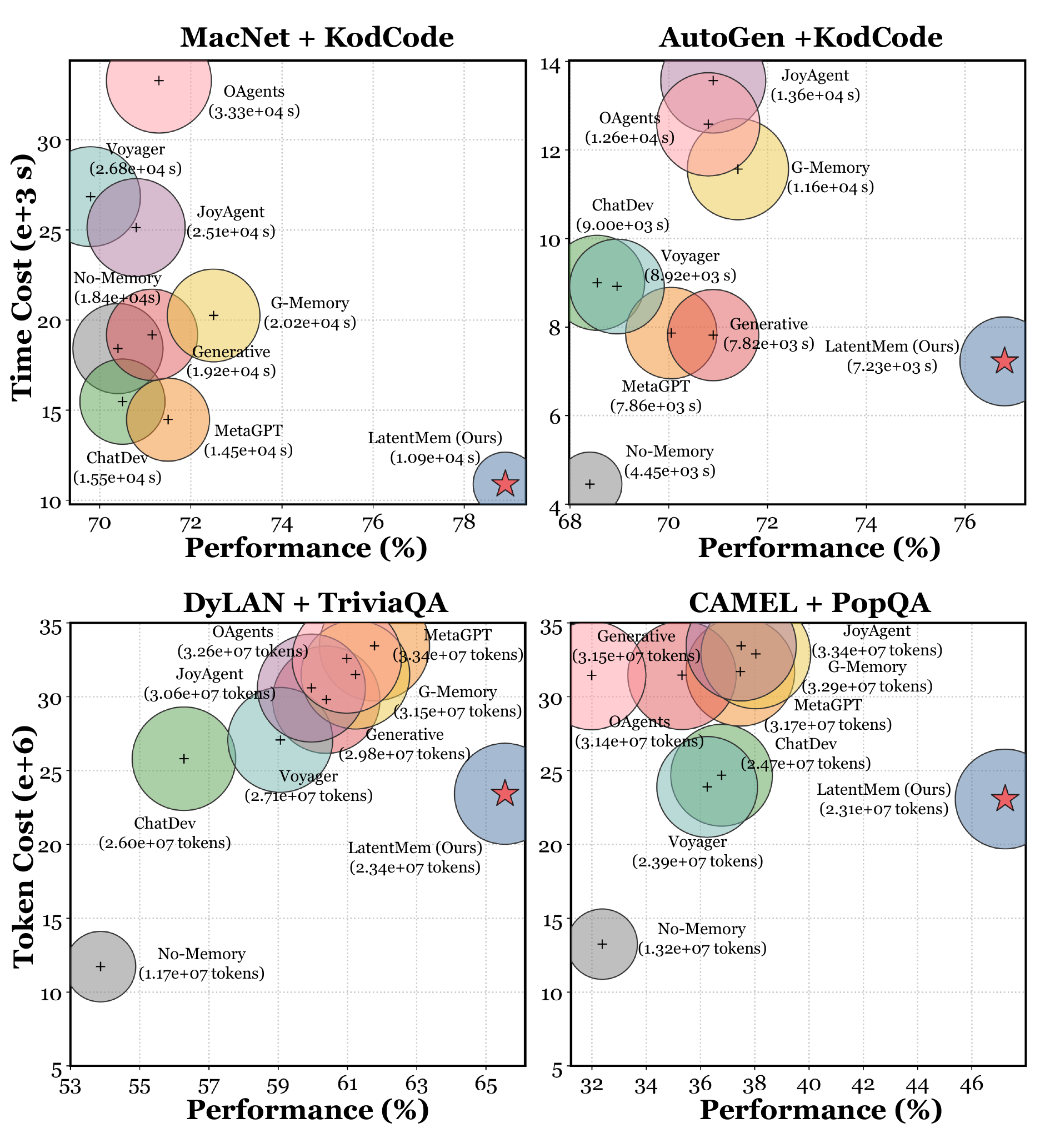}
    \caption{Time and token consumption of LatentMem. Each panel shows the trade-off between performance and resource cost under different memory frameworks: the top row plots performance versus time, the bottom row plots performance versus token cost. Circle area reflects relative resource consumption. “A + B” indicates MAS framework A evaluated on dataset B.}
    \label{fig:app_cost}
\end{figure}

\Cref{fig:app_cost} provides additional comparisons of token and time costs across various benchmarks and MAS frameworks with different memory architectures. Overall, \ourmethod leverages its latent memory design to achieve the fastest inference and lowest token usage among all memory baselines, while consistently delivering the most substantial improvements.

\subsection{Role-aware Memory Analysis}\label{app:role_aware}
\begin{figure}[!t]
    \centering
    \includegraphics[width=1\linewidth]{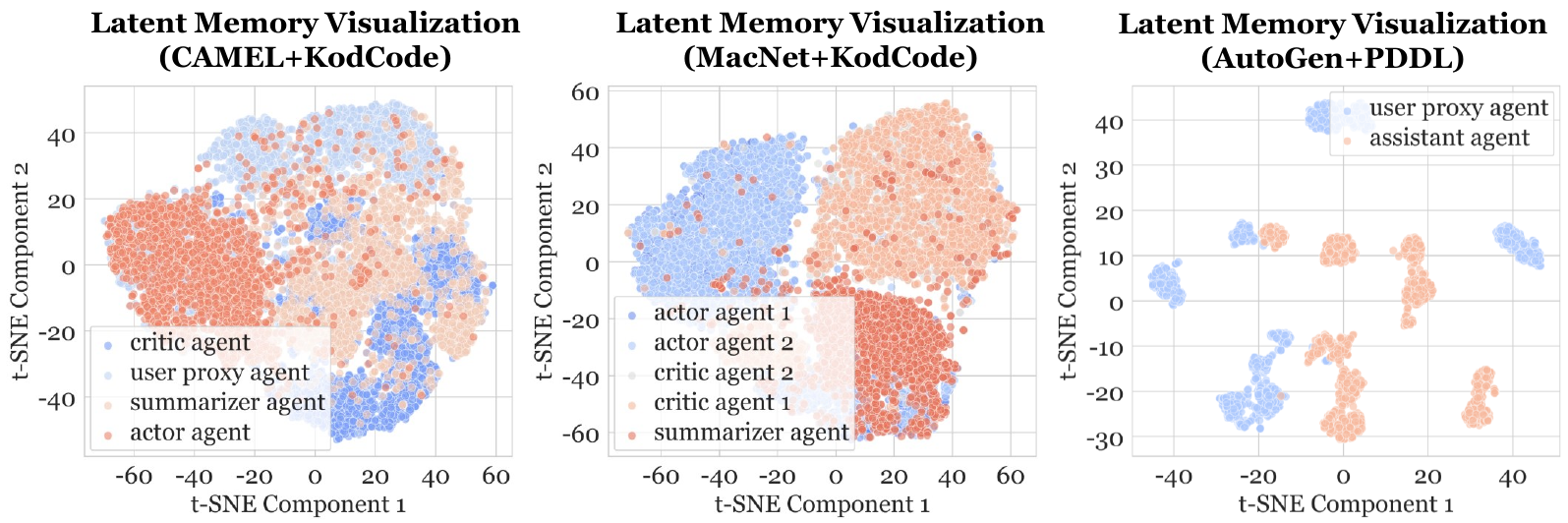}
    \caption{t-SNE visualization of latent memories generated by \ourmethod across different datasets and MAS frameworks.}
    \label{fig:app_role}
\end{figure}

This section outlines the approach used to examine the geometric organization of latent memories (illustrated in \Cref{fig:role,fig:app_role}) and to discover semantically meaningful clusters. The workflow consists of two main steps: first, deriving a fixed-size vector representation for each memory sequence, and second, applying dimensionality reduction and clustering techniques.

Consider $\{m_1, m_2, \dots, m_B\}$ as a collection of $B$ latent memory sequences
obtained from \ourmethod's inference, where each sequence 
$m_i = (m_{i,1}, \dots, m_{i,L'}) \in \mathbb{R}^{L' \times D}$ 
contains $L'$ token embeddings of dimensionality $D$.

To generate a single, compact representation for each sequence,
we compute its mean embedding $\bar{m}_i \in \mathbb{R}^{D}$ as follows:
\begin{equation}
\bar{m}_i = \frac{1}{L'} \sum_{l=1}^{L'} m_{i,l}.
\end{equation}

The resulting set of high-dimensional vectors, $\{\bar{m}_1, \bar{m}_2, \dots, \bar{m}_B\}$, is then used as input for downstream analysis. 

For visualization purposes, we apply t-SNE~\citep{maaten2008visualizing} to map the vectors 
$\{\bar{m}_1, \bar{m}_2, \dots, \bar{m}_B\}$ into a two-dimensional space. Formally, this transformation is defined as
\begin{equation}
y_i \in \mathbb{R}^2 = f_{\text{t-SNE}}(\bar{m}_i).
\end{equation}

These 2D embeddings are subsequently employed to produce the scatter plots.

\subsection{Sensitivity Analysis}\label{app:sensitivity}
\begin{figure}[!t]
    \centering
    \includegraphics[width=0.6\linewidth]{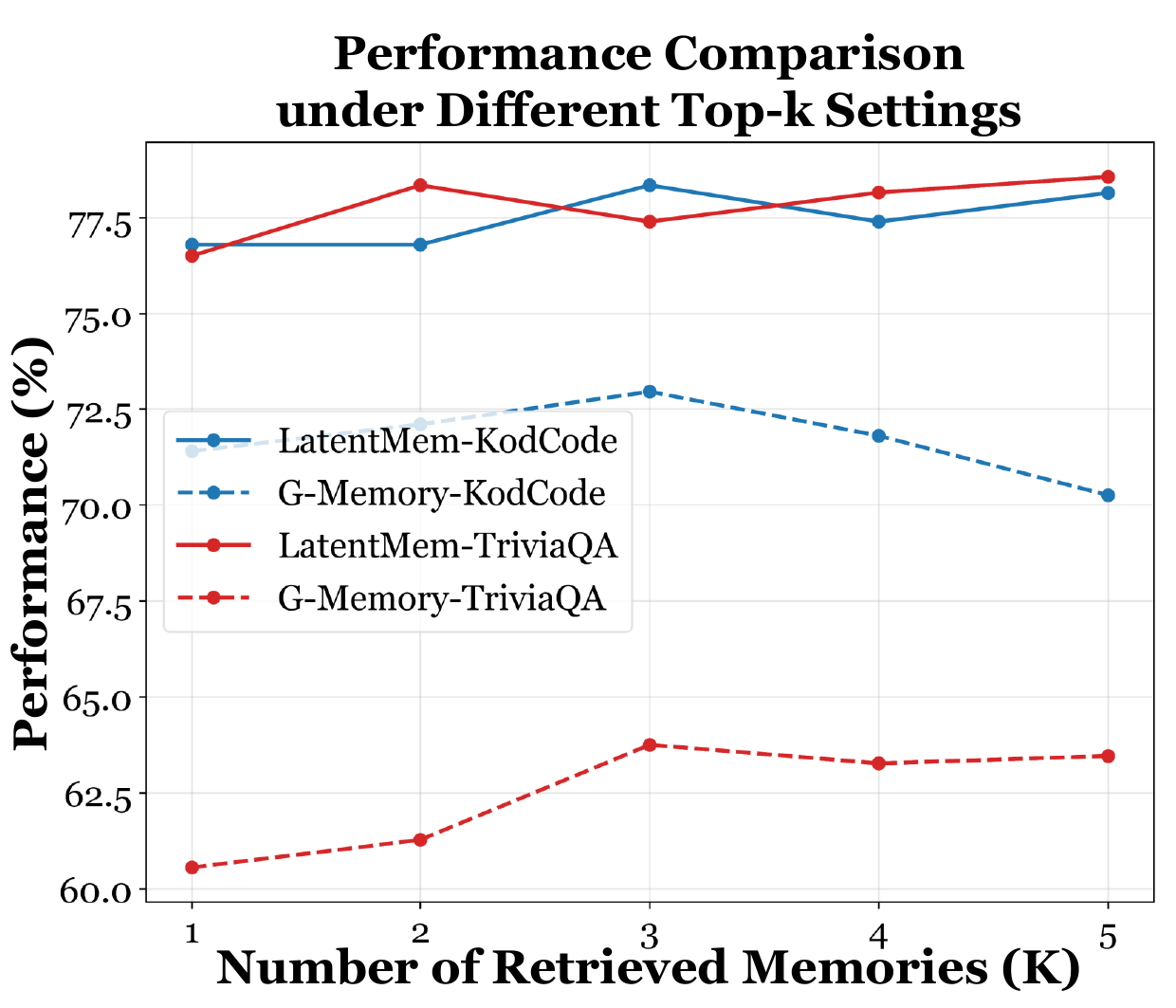}
    \caption{Comparison between \ourmethod and G-Memory under varying top-$K$ settings on KodCode and TriviaQA with AutoGen. While \ourmethod consistently benefits from more trajectories as $K$ increases, G-Memory's performance declines when $K>3$.}
    \label{fig:topk}
\end{figure}

As a supplement to \Cref{sec:ablation}, we conduct a sensitivity analysis of \ourmethod with respect to the top-$K$ raw trajectories retrieved from the experience bank, as shown in \Cref{fig:topk}. It can be observed that G-Memory initially benefits from increasing $K$, but its performance starts to decline when $K>3$; for example, on KodCode, accuracy drops from 72.95\% at $K=3$ to 70.25\% at $K=5$. In contrast, \ourmethod continues to improve as $K$ increases, highlighting its ability to leverage fixed-length, role-aware latent memory to distill useful information even from redundant trajectories.

\begin{table*}[!t]
\caption{Performance comparison with diverse memory frameworks on six benchmarks. The underlying LLM backbone is \llmname{Llama-3.1-8B-Instruct}. We highlight the \hlfirst{best} and \hlsecond{second best} results. TriviaQA, KodCode, StrategyQA, and PopQA are in-domain benchmarks, while BigCodeBench and PDDL are out-of-domain; AutoGen and MacNet are in-distribution MAS frameworks, whereas CAMEL and DyLAN are unseen frameworks. Additional details are provided in Appendix \ref{app:train_details}.
}
\label{tab:results_8b}

\renewcommand{\arraystretch}{1.1}
\centering \scriptsize \setlength{\tabcolsep}{8pt}

\resizebox{\textwidth}{!}{
\begin{tabular}{rl|l|cccc|cc|c}
\toprule
\multicolumn{2}{c|}{\multirow{2}{*}{}} & \multirow{2}{*}{\textbf{Method}} & \multicolumn{4}{c|}{\textbf{Held-in}} & \multicolumn{2}{c|}{\textbf{Held-out}} & \multirow{2}{*}{\textbf{Average}} \\
\multicolumn{2}{c|}{} & & \textbf{TriviaQA} & \textbf{KodCode}  & \textbf{StrategyQA} & \textbf{PopQA} & \textbf{BigCodeBench} & \textbf{PDDL} \\
\midrule

\multirow{14}{*}{\rotatebox[origin=c]{90}{\textbf{Held-in}}\hspace{-3em}} & \multirow{7}{*}{\rotatebox[origin=c]{90}{\textbf{AutoGen}}} 
& No-memory & 72.03\upgrade{0.00} & 47.45\upgrade{0.00} & 61.50\upgrade{0.00} & 41.20\upgrade{0.00} & 80.96\upgrade{0.00} & 14.03\upgrade{0.00} & 52.86\upgrade{0.00} \\
& & MetaGPT & 74.10\upgrade{2.07} & 50.00\upgrade{2.55} & 58.73\degrade{2.77} & \hlsecond{47.41\upgrade{6.21}} & 82.11\upgrade{1.15} & \hlsecond{25.61\upgrade{11.58}} & 56.33\upgrade{3.47} \\
& & Generative & 73.77\upgrade{1.74} & 49.40\upgrade{1.95} & 63.97\upgrade{2.47} & 46.00\upgrade{4.80} & 82.11\upgrade{1.15} & 20.63\upgrade{6.60} & 55.98\upgrade{3.12} \\
& & Voyager & 74.53\upgrade{2.50} & 46.45\degrade{1.00} & 61.28\degrade{0.22} & 46.75\upgrade{5.55} & 81.49\upgrade{0.53} & 21.65\upgrade{7.62} & 55.36\upgrade{2.50} \\
& & G-Memory & \hlsecond{74.60\upgrade{2.57}} & \hlsecond{50.20\upgrade{2.75}} & 64.20\upgrade{2.70} & 47.24\upgrade{6.04} & \hlsecond{82.67\upgrade{1.71}} & 24.31\upgrade{10.28} & \hlsecond{57.20\upgrade{4.34}} \\
& & OAgent & 73.12\upgrade{1.09} & 49.55\upgrade{2.10} & \hlsecond{65.94\upgrade{4.44}} & 47.25\upgrade{6.05} & 82.11\upgrade{1.15} & 16.99\upgrade{2.96} & 55.83\upgrade{2.97} \\
& & \ourmethod & \hlfirst{74.92\upgrade{2.89}} & \hlfirst{65.90\upgrade{18.45}} & \hlfirst{67.89\upgrade{6.39}} & \hlfirst{49.40\upgrade{8.20}} & \hlfirst{83.84\upgrade{2.88}} & \hlfirst{28.96\upgrade{14.93}} & \hlfirst{61.82\upgrade{8.96}} \\
\cmidrule{3-10}

& \multirow{7}{*}{\rotatebox[origin=c]{90}{\textbf{MacNet}}} 
& No-memory & 71.62\upgrade{0.00} & 48.50\upgrade{0.00} & 62.94\upgrade{0.00} & 40.59\upgrade{0.00} & 79.28\upgrade{0.00} & 17.33\upgrade{0.00} & 53.38\upgrade{0.00} \\
& & MetaGPT & 72.57\upgrade{0.95} & \hlsecond{50.30\upgrade{1.80}} & 62.66\degrade{0.28} & 48.42\upgrade{7.83} & 81.28\upgrade{2.00} & 21.47\upgrade{4.14} & 56.12\upgrade{2.74} \\
& & Generative & \hlsecond{72.83\upgrade{1.21}} & 49.85\upgrade{1.35} & 60.80\degrade{2.14} & 47.32\upgrade{6.73} & 81.11\upgrade{1.83} & \hlsecond{24.80\upgrade{7.47}} & \hlsecond{56.12\upgrade{2.74}} \\
& & Voyager & 72.17\upgrade{0.55} & 44.80\degrade{3.70} & 61.35\degrade{1.59} & 47.11\upgrade{6.52} & 81.67\upgrade{2.39} & 18.97\upgrade{1.64} & 54.35\upgrade{0.97} \\
& & G-Memory & 71.79\upgrade{0.17} & 49.20\upgrade{0.70} & 62.23\degrade{0.71} & \hlsecond{48.96\upgrade{8.37}} & 81.46\upgrade{2.18} & 19.10\upgrade{1.77} & 55.46\upgrade{2.08} \\
& & OAgent & 72.20\upgrade{0.58} & 48.50\upgrade{0.00} & \hlsecond{63.76\upgrade{0.82}} & 48.30\upgrade{7.71} & \hlsecond{82.28\upgrade{3.00}} & 16.71\degrade{0.62} & 55.29\upgrade{1.91} \\
& & \ourmethod & \hlfirst{74.45\upgrade{2.83}} & \hlfirst{65.50\upgrade{17.00}} & \hlfirst{66.42\upgrade{3.48}} & \hlfirst{50.16\upgrade{9.57}} & \hlfirst{83.84\upgrade{4.56}} & \hlfirst{25.61\upgrade{8.28}} & \hlfirst{61.00\upgrade{7.62}} \\
\midrule

\multirow{14}{*}{\rotatebox[origin=c]{90}{\textbf{Held-out}}\hspace{-3em}} & \multirow{7}{*}{\rotatebox[origin=c]{90}{\textbf{CAMEL}}} 
& No-memory & 72.86\upgrade{0.00} & 48.95\upgrade{0.00} & 60.25\upgrade{0.00} & 40.44\upgrade{0.00} & 80.28\upgrade{0.00} & 17.45\upgrade{0.00} & 53.37\upgrade{0.00} \\
& & MetaGPT & 72.99\upgrade{0.13} & 51.70\upgrade{2.75} & \hlsecond{62.66\upgrade{2.41}} & 47.68\upgrade{7.24} & 81.93\upgrade{1.65} & 23.24\upgrade{5.79} & \hlsecond{56.70\upgrade{3.33}} \\
& & Generative & 72.36\degrade{0.50} & 49.80\upgrade{0.85} & 60.26\upgrade{0.01} & \hlsecond{49.09\upgrade{8.65}} & 80.28\upgrade{0.00} & \hlsecond{23.97\upgrade{6.52}} & 55.96\upgrade{2.59} \\
& & Voyager & 72.57\degrade{0.29} & 48.95\upgrade{0.00} & 61.57\upgrade{1.32} & 48.68\upgrade{8.24} & 81.75\upgrade{1.47} & 15.79\degrade{1.66} & 54.88\upgrade{1.51} \\
& & G-Memory & \hlsecond{73.69\upgrade{0.83}} & 49.50\upgrade{0.55} & 61.14\upgrade{0.89} & 48.80\upgrade{8.36} & \hlsecond{81.93\upgrade{1.65}} & 20.69\upgrade{3.24} & 55.96\upgrade{2.59} \\
& & OAgent & 73.52\upgrade{0.66} & \hlsecond{49.90\upgrade{0.95}} & 61.48\upgrade{1.23} & 47.90\upgrade{7.46} & 81.49\upgrade{1.21} & 17.12\degrade{0.33} & 55.23\upgrade{1.86} \\
& & \ourmethod & \hlfirst{74.00\upgrade{1.14}} & \hlfirst{63.85\upgrade{14.90}} & \hlfirst{65.28\upgrade{5.03}} & \hlfirst{49.90\upgrade{9.46}} & \hlfirst{82.75\upgrade{2.47}} & \hlfirst{24.46\upgrade{7.01}} & \hlfirst{60.04\upgrade{6.67}} \\
\cmidrule{3-10}

& \multirow{7}{*}{\textbf{\rotatebox[origin=c]{90}{\textbf{DyLAN}}}} 
& No-memory & 72.62\upgrade{0.00} & 47.55\upgrade{0.00} & 60.67\upgrade{0.00} & 40.50\upgrade{0.00} & 80.26\upgrade{0.00} & 15.20\upgrade{0.00} & 52.80\upgrade{0.00} \\
& & MetaGPT & 73.87\upgrade{1.25} & \hlsecond{49.85\upgrade{2.30}} & 62.88\upgrade{2.21} & \hlsecond{47.92\upgrade{7.42}} & 81.28\upgrade{1.02} & 18.45\upgrade{3.25} & \hlsecond{55.71\upgrade{2.91}} \\
& & Generative & 73.22\upgrade{0.60} & 47.10\degrade{0.45} & \hlsecond{64.85\upgrade{4.18}} & 48.55\upgrade{8.05} & 82.19\upgrade{1.93} & 17.69\upgrade{2.49} & 55.60\upgrade{2.80} \\
& & Voyager & 73.69\upgrade{1.07} & 44.80\degrade{2.75} & 61.35\upgrade{0.68} & 48.88\upgrade{8.38} & 81.58\upgrade{1.32} & 14.33\degrade{0.87} & 54.10\upgrade{1.30} \\
& & G-Memory & \hlsecond{74.00\upgrade{1.38}} & 48.80\upgrade{1.25} & 61.35\upgrade{0.68} & 47.32\upgrade{6.82} & \hlsecond{82.28\upgrade{2.02}} & \hlsecond{19.64\upgrade{4.44}} & 55.56\upgrade{2.76} \\
& & OAgent & 73.49\upgrade{0.87} & 48.25\upgrade{0.70} & 63.25\upgrade{2.58} & 47.90\upgrade{7.40} & 81.30\upgrade{1.04} & 16.75\upgrade{1.55} & 55.16\upgrade{2.36} \\
& & \ourmethod & \hlfirst{74.10\upgrade{1.48}} & \hlfirst{64.25\upgrade{16.70}} & \hlfirst{66.83\upgrade{6.16}} & \hlfirst{49.34\upgrade{8.84}} & \hlfirst{83.25\upgrade{2.99}} & \hlfirst{22.62\upgrade{7.42}} & \hlfirst{60.06\upgrade{7.26}} \\
\bottomrule

\end{tabular}
}

\end{table*}

\subsection{Case Study}\label{app:case_study}
\Cref{fig:case_study} illustrates how \ourmethod, by providing role-aware memory, can prevent or promptly correct common error patterns in MAS. For example, when solving the PDDL task: \textit{``Satisfy the following conditions: ball1 is at room b, ball2 is at room a, ball3 is at room b, ball4 is at room b, ball5 is at room a, ball6 is at room a''}, the vanilla MacNet repeatedly moves ball2 between room a and room b while ignoring the task’s termination conditions, resulting in a loop, which is called \textit{Step Repetition} error. MacNet with OAgents, although able to retrieve similar historical trajectories, is influenced by excessively long histories and fails to account for differences between the current task and the retrieved trajectories. This leads to blindly following the retrieved steps without satisfying the actual task requirements, which is called \textit{Disobey Task Specification} error. Moreover, both methods suffer from significant \textit{Reasoning-Action Mismatch}, where the agent’s planned reasoning steps are correct, but its executed actions deviate from the intended plan. 

In contrast, \ourmethod provides high-level, role-aware latent memory to each agent at every reasoning step. By leveraging personalized high-level historical experience, each agent naturally reinforces its role compliance, thereby enabling the actor-critic style coordination mechanism in MacNet to function effectively. Even when a short-term error occurs in the trajectory (e.g., generating an incorrect action "pick ball5 rooma right"), the MAS can immediately self-correct in the next reasoning step using a ``check valid answer'' procedure, ultimately completing the task successfully.

\section{Prompt Templates}
In this section, we provide the prompt templates used in our framework. Note that these templates may vary slightly across different task domains. As an example, we present the prompt template of CAMEL and AutoGen for code tasks.

\begin{tcolorbox}[notitle, sharp corners, breakable, colframe=Periwinkle, colback=white, 
       boxrule=3pt, boxsep=0.5pt, enhanced, 
       shadow={3pt}{-3pt}{0pt}{opacity=1,mygrey},
       title={CAMEL's Prompt Template for Strategy Agent},]
       {\scriptsize
\begin{lstlisting}[
    language={},
    basicstyle=\ttfamily\scriptsize,
    breaklines=true,
    columns=fullflexible,
    numbers=none,
    escapeinside={(*@}{@*)},
]
You are a strategy-generation agent. Your task is to read a given coding problem and provide a **detailed implementation strategy**, but **do not write any code**.

# Objectives
- Understand the problem requirements.
- Describe the algorithm, data structures, and step-by-step approach.
- Ensure the strategy is clear enough for a developer tao implement directly.

# Output Guidelines
- Focus on logic and process; avoid including actual code or irrelevant explanations.
- You should keep your response concise, no more than 3 sentences.
\end{lstlisting}
}
\end{tcolorbox}

\begin{tcolorbox}[notitle, sharp corners, breakable, colframe=Periwinkle, colback=white, 
       boxrule=3pt, boxsep=0.5pt, enhanced, 
       shadow={3pt}{-3pt}{0pt}{opacity=1,mygrey},
       title={CAMEL's Prompt Template for Code Agent},]
       {\scriptsize
\begin{lstlisting}[
    language={},
    basicstyle=\ttfamily\scriptsize,
    breaklines=true,
    columns=fullflexible,
    numbers=none,
    escapeinside={(*@}{@*)},
]
You are a Code Implementation agent. You will be provided with a problem and an analysis of that problem from a user agent. Your task is to produce complete and correct code implementations based on coding problems.

# Objectives
- Write clear, well-structured, and correct Python code.
- Do not include any explanations or comments outside the code.

# Output Guidelines
- Wrap the entire Python code inside a code block using triple backticks:
```python
# your code here
```
\end{lstlisting}
}
\end{tcolorbox}

\begin{tcolorbox}[notitle, sharp corners, breakable, colframe=Periwinkle, colback=white, 
       boxrule=3pt, boxsep=0.5pt, enhanced, 
       shadow={3pt}{-3pt}{0pt}{opacity=1,mygrey},
       title={CAMEL's Prompt Template for Test Agent},]
       {\scriptsize
\begin{lstlisting}[
    language={},
    basicstyle=\ttfamily\scriptsize,
    breaklines=true,
    columns=fullflexible,
    numbers=none,
    escapeinside={(*@}{@*)},
]
You are a code evaluator. Your task is to review the current coding problem and the code written by the actor agent for that problem.

- If the code is correct, reply only with: "Agree".
- If the code has issues, give brief and concise feedback only(Keep your response short and within 3 sentences).
\end{lstlisting}
}
\end{tcolorbox}

\begin{tcolorbox}[notitle, sharp corners, breakable, colframe=Periwinkle, colback=white, 
       boxrule=3pt, boxsep=0.5pt, enhanced, 
       shadow={3pt}{-3pt}{0pt}{opacity=1,mygrey},
       title={CAMEL's Prompt Template for Summarizer Agent},]
       {\scriptsize
\begin{lstlisting}[
    language={},
    basicstyle=\ttfamily\scriptsize,
    breaklines=true,
    columns=fullflexible,
    numbers=none,
    escapeinside={(*@}{@*)},
]
You are a summarization and final-code-generation agent. Your task is to read the previous actor code implementations and the corresponding critic improvement suggestions, and then produce the final, corrected, and consolidated code solution for the current task.

# Objectives
- Carefully examine the actor's code solutions.
- Incorporate the critic's improvement suggestions when necessary.
- Produce a clean, complete, and correct final code implementation.
- Do not include explanations, comments, or any text outside the code block.

# Output Format
- Wrap the entire final Python code inside triple backticks:
```python
# final code here
```
\end{lstlisting}
}
\end{tcolorbox}

\begin{tcolorbox}[notitle, sharp corners, breakable, colframe=Periwinkle, colback=white, 
       boxrule=3pt, boxsep=0.5pt, enhanced, 
       shadow={3pt}{-3pt}{0pt}{opacity=1,mygrey},
       title={AutoGen's Prompt Template for Assistant Agent},]
       {\scriptsize
\begin{lstlisting}[
    language={},
    basicstyle=\ttfamily\scriptsize,
    breaklines=true,
    columns=fullflexible,
    numbers=none,
    escapeinside={(*@}{@*)},
]
You are a strategy-generation agent. Your task is to read a given coding problem and provide a **detailed implementation strategy**, but **do not write any code**.

# Objectives
- Understand the problem requirements.
- Describe the algorithm, data structures, and step-by-step approach.
- Ensure the strategy is clear enough for a developer tao implement directly.

# Output Guidelines
- Focus on logic and process; avoid including actual code or irrelevant explanations.
- You should keep your response concise, no more than 3 sentences.
\end{lstlisting}
}
\end{tcolorbox}

\begin{tcolorbox}[notitle, sharp corners, breakable, colframe=Periwinkle, colback=white, 
       boxrule=3pt, boxsep=0.5pt, enhanced, 
       shadow={3pt}{-3pt}{0pt}{opacity=1,mygrey},
       title={AutoGen's Prompt Template for User Proxy Agent},]
       {\scriptsize
\begin{lstlisting}[
    language={},
    basicstyle=\ttfamily\scriptsize,
    breaklines=true,
    columns=fullflexible,
    numbers=none,
    escapeinside={(*@}{@*)},
]
You are a Code Implementation agent. Your task is to read the implementation strategy provided by the Assistant agent and produce **complete, executable Python code** that follows the strategy exactly.

# Objectives
- Implement the solution according to the detailed strategy from the Assistant.
- Write clear, well-structured, and correct Python code.
- Make sure the code covers all steps and handles edge cases mentioned in the strategy.
- Do not include any explanations or comments outside the code.

# Output Guidelines
- Wrap the entire Python code inside a code block using triple backticks:
```python
# your code here
```
\end{lstlisting}
}
\end{tcolorbox}

\section{Example}
In this section, we present example trajectories generated by \ourmethod-enhanced AutoGen on Knowledge QA (TriviaQA) and Game (PDDL) tasks to demonstrate the effectiveness of \ourmethod.
\begin{tcolorbox}[notitle, sharp corners, breakable, colframe=Periwinkle, colback=white, 
       boxrule=3pt, boxsep=0.5pt, enhanced, 
       shadow={3pt}{-3pt}{0pt}{opacity=1,mygrey},
       title={Example trajectory on PDDL},]
       {\scriptsize
\begin{lstlisting}[
    language={},
    basicstyle=\ttfamily\scriptsize,
    breaklines=true,
    columns=fullflexible,
    numbers=none,
    escapeinside={(*@}{@*)},
]
<|im_start|>user
## Successful Examples (Reference Cases)
Below are some examples of similar tasks that were successfully completed.  
Please use these as references to guide your thinking and approach to the current task:

<|im_start|>user
<observation>Ball1 is a ball. Ball1 is at rooma. Ball2 is a ball. Ball2 is at rooma. Ball3 is a ball. Ball3 is at rooma. Left is a gripper. Left is free. Right is a gripper. Right is free. Robby is at rooma. Room rooma Room roomb</observation><|im_end|>
<|im_start|>assistant
<action>Pick up ball1 at rooma with arm right.</action><|im_end|>
<|im_start|>user
<observation>Ball1 is a ball. Ball1 is carrying right. Ball2 is a ball. Ball2 is at rooma. Ball3 is a ball. Ball3 is at rooma.  Left is a gripper. Left is free. Right is a gripper. Robby is at rooma. Room rooma Room roomb</observation><|im_end|>
<|im_start|>assistant
<action>Pick up ball2 at rooma with arm left.</action><|im_end|>
<|im_start|>user
<observation>Ball1 is a ball. Ball1 is carrying right. Ball2 is a ball. Ball2 is carrying left. Ball3 is a ball. Ball3 is at rooma. Left is a gripper. Right is a gripper. Robby is at rooma. Room rooma Room roomb</observation><|im_end|>
<|im_start|>assistant
<action>Move from rooma to roomb</action><|im_end|>
<|im_start|>user
<observation>Ball1 is a ball. Ball1 is carrying right. Ball2 is a ball. Ball2 is carrying left. Ball3 is a ball. Ball3 is at rooma. Left is a gripper. Right is a gripper. Robby is at roomb. Room rooma Room roomb</observation><|im_end|>
<|im_start|>assistant
<action>drop ball1 at rooma with arm right</action><|im_end|>
<|im_start|>user
<observation>The action is not valid and therefore takes no effect. Please remember to satisfy the restriction of actions. You can also check valid actions.</observation><|im_end|>
<|im_start|>assistant
<action>check valid actions</action><|im_end|>
<|im_start|>user
<observation>Valid actions are: drop ball1 roomb right, drop ball2 roomb left, Move from roomb to rooma. , check valid actions</observation><|im_end|>
<|im_start|>assistant
<action>drop ball1 roomb right</action><|im_end|>
<|im_start|>user
<observation>Ball1 is a ball. Ball1 is at roomb. Ball2 is a ball. Ball2 is carrying left. Ball3 is a ball. Ball3 is at rooma. Left is a gripper. Right is a gripper. Right is free. Robby is at roomb. Room rooma Room roomb</observation><|im_end|>
<|im_start|>assistant
<action>drop ball2 roomb left</action><|im_end|>
<|im_start|>user
<observation>Ball1 is a ball. Ball1 is at roomb. Ball2 is a ball. Ball2 is at roomb. Ball3 is a ball. Ball3 is at rooma. Left is a gripper. Left is free. Right is a gripper. Right is free. Robby is at roomb. Room rooma Room roomb</observation><|im_end|>
<|im_start|>assistant
<action>move from roomb to rooma</action><|im_end|>
<|im_start|>user
<observation>Ball1 is a ball. Ball1 is at roomb. Ball2 is a ball. Ball2 is at roomb. Ball3 is a ball. Ball3 is at rooma.Left is a gripper. Left is free. Right is a gripper. Right is free. Robby is at rooma. Room rooma Room roomb</observation><|im_end|>
<|im_start|>assistant
<action>Pick up ball3 at rooma with arm right.</action><|im_end|>
<|im_start|>user
<observation>Ball1 is a ball. Ball1 is at roomb. Ball2 is a ball. Ball2 is at roomb. Ball3 is a ball. Ball3 is carrying right.  Left is a gripper. Left is free. Right is a gripper. Robby is at rooma. Room rooma Room roomb</observation><|im_end|>
<|im_start|>assistant
<action>Move from rooma to roomb</action><|im_end|>
<|im_start|>user
<observation>Ball1 is a ball. Ball1 is at roomb. Ball2 is a ball. Ball2 is at roomb. Ball3 is a ball. Ball3 is carrying right. Ball4 is a ball. Ball4 is at rooma. Left is a gripper. Left is free. Right is a gripper. Robby is at roomb. Room rooma Room roomb</observation><|im_end|>
<|im_start|>assistant
<action>drop ball3 rooms right</action><|im_end|>
<|im_start|>user
<observation>Ball1 is a ball. Ball1 is at roomb. Ball2 is a ball. Ball2 is at roomb. Ball3 is a ball. Ball3 is at roomb. Ball4 is a ball. Ball4 is at rooma. Left is a gripper. Left is free. Right is a gripper. Right is free. Robby is at roomb. Room rooma Room roomb. The task is completed.</observation><|im_end|>
---

Use the above examples and insights as a foundation, and now work on the following task:

You are a robot with a gripper that can move objects between different rooms. Your name is Robby.

## There are three actions defined in this domain:
- think xxx: 
  - Format: `think: I need to move block A first because...`  
  - Any other format is invalid and will not be recognized by the system.
- move <room1> <room2>: This action allows the robot to move from one room to another.The action has a single precondition, which is that the robot is currently in a room. The effect of this action is to move the robot to another room and to remove the fact that it is in the original room.
- pick <obj> <room> <gripper>: This action allows the robot to pick up an object using the gripper. The action has three preconditions: (1) the object is located in a room (2) the robot is currently in the same room and (3) the gripper is free (i.e., not holding any object). The effect of this action is to update the state of the world to show that the robot is carrying the object using the gripper, the object is no longer in the room, and the gripper is no longer free.
- drop <obj> <room> <gripper>: This action allows the robot to drop an object that it is carrying. The action has two preconditions: (1) the robot is currently carrying the object using the gripper, and (2) the robot is currently in a room. The effect of this action is to update the state of the world to show that the robot is no longer carrying the object using the gripper, the object is now located in the room, and the gripper is now free.

## NOTE: 
- You should strictly follow these actions, no any other actions are allowed!
- Before you respond, you must ensure that your output actions strictly comply with the rules described above!

Here is your initial observation: Ball1 is a ball.  Ball1 is at rooma.  Ball2 is a ball.  Ball2 is at rooma.  Ball3 is a ball.  Ball3 is at roomb.  Ball4 is a ball.  Ball4 is at roomb.  Left is a gripper.  Left is free.  Right is a gripper.  Right is free.  Robby is at rooma.  Room rooma Room roomb
**Here is your task: The goal is to satisfy the following conditions: ball1 is at roomb. , ball2 is at roomb. , ball3 is at rooma. , ball4 is at rooma.<|im_end|>
<|im_start|>assistant
<action>pick ball1 rooma right</action><|im_end|>
<|im_start|>user
Ball1 is a ball.  Ball1 is carrying right.  Ball2 is a ball.  Ball2 is at rooma.  Ball3 is a ball.  Ball3 is at roomb.  Ball4 is a ball.  Ball4 is at roomb.  Left is a gripper.  Left is free.  Right is a gripper.  Robby is at rooma.  Room rooma Room roomb<|im_end|>
<|im_start|>assistant
<action>move from rooma to roomb</action><|im_end|>
<|im_start|>user
Ball1 is a ball.  Ball1 is carrying right.  Ball2 is a ball.  Ball2 is at rooma.  Ball3 is a ball.  Ball3 is at roomb.  Ball4 is a ball.  Ball4 is at roomb.  Left is a gripper.  Left is free.  Right is a gripper.  Robby is at roomb.  Room rooma Room roomb<|im_end|>
<|im_start|>assistant
<action>drop ball1 roomb right</action><|im_end|>
<|im_start|>user
Ball1 is a ball.  Ball1 is at roomb.  Ball2 is a ball.  Ball2 is at rooma.  Ball3 is a ball.  Ball3 is at roomb.  Ball4 is a ball.  Ball4 is at roomb.  Left is a gripper.  Left is free.  Right is a gripper.  Right is free.  Robby is at roomb.  Room rooma Room roomb<|im_end|>
<|im_start|>assistant
<action>pick ball2 rooma left</action><|im_end|>
<|im_start|>user
The action is not valid and therefore takes no effect. You should use `check valid actions.` command to get some clues!<|im_end|>
<|im_start|>assistant
<action>check valid actions</action><|im_end|>
<|im_start|>user
Valid actions are: Pick up ball1 at roomb with arm left. , Pick up ball3 at roomb with arm left. , Move from roomb to rooma. , Pick up ball3 at roomb with arm right. , Pick up ball4 at roomb with arm left. , Pick up ball1 at roomb with arm right. , Pick up ball4 at roomb with arm right. , check valid actions<|im_end|>
<|im_start|>assistant
<action>pick ball3 roomb left</action><|im_end|>
<|im_start|>user
Ball1 is a ball.  Ball1 is at roomb.  Ball2 is a ball.  Ball2 is at rooma.  Ball3 is a ball.  Ball3 is carrying left.  Ball4 is a ball.  Ball4 is at roomb.  Left is a gripper.  Right is a gripper.  Right is free.  Robby is at roomb.  Room rooma Room roomb<|im_end|>
<|im_start|>assistant
<action>move from roomb to rooma</action><|im_end|>
<|im_start|>user
Ball1 is a ball.  Ball1 is at roomb.  Ball2 is a ball.  Ball2 is at rooma.  Ball3 is a ball.  Ball3 is carrying left.  Ball4 is a ball.  Ball4 is at roomb.  Left is a gripper.  Right is a gripper.  Right is free.  Robby is at rooma.  Room rooma Room roomb<|im_end|>
<|im_start|>assistant
<action>drop ball3 rooma left</action><|im_end|>
<|im_start|>user
Ball1 is a ball.  Ball1 is at roomb.  Ball2 is a ball.  Ball2 is at rooma.  Ball3 is a ball.  Ball3 is at rooma.  Ball4 is a ball.  Ball4 is at roomb.  Left is a gripper.  Left is free.  Right is a gripper.  Right is free.  Robby is at rooma.  Room rooma Room roomb<|im_end|>
<|im_start|>assistant
<action>pick ball4 roomb right</action><|im_end|>
<|im_start|>user
The action is not valid and therefore takes no effect. You should use `check valid actions.` command to get some clues!<|im_end|>
<|im_start|>assistant
<action>check valid actions</action><|im_end|>
<|im_start|>user
Valid actions are: Pick up ball2 at rooma with arm right. , Pick up ball3 at rooma with arm right. , Pick up ball3 at rooma with arm left. , Pick up ball2 at rooma with arm left. , Move from rooma to roomb. , check valid actions<|im_end|>
<|im_start|>assistant
<action>pick ball2 rooma right</action><|im_end|>
<|im_start|>user
Ball1 is a ball.  Ball1 is at roomb.  Ball2 is a ball.  Ball2 is carrying right.  Ball3 is a ball.  Ball3 is at rooma.  Ball4 is a ball.  Ball4 is at roomb.  Left is a gripper.  Left is free.  Right is a gripper.  Robby is at rooma.  Room rooma Room roomb<|im_end|>
<|im_start|>assistant
<action>move from rooma to roomb</action><|im_end|>
<|im_start|>user
Ball1 is a ball.  Ball1 is at roomb.  Ball2 is a ball.  Ball2 is carrying right.  Ball3 is a ball.  Ball3 is at rooma.  Ball4 is a ball.  Ball4 is at roomb.  Left is a gripper.  Left is free.  Right is a gripper.  Robby is at roomb.  Room rooma Room roomb<|im_end|>
<|im_start|>assistant
<action>drop ball2 roomb right</action><|im_end|>
<|im_start|>user
Ball1 is a ball.  Ball1 is at roomb.  Ball2 is a ball.  Ball2 is at roomb.  Ball3 is a ball.  Ball3 is at rooma.  Ball4 is a ball.  Ball4 is at roomb.  Left is a gripper.  Left is free.  Right is a gripper.  Right is free.  Robby is at roomb.  Room rooma Room roomb<|im_end|>
<|im_start|>assistant
<action>pick ball4 roomb right</action><|im_end|>
<|im_start|>user
Ball1 is a ball.  Ball1 is at roomb.  Ball2 is a ball.  Ball2 is at roomb.  Ball3 is a ball.  Ball3 is at rooma.  Ball4 is a ball.  Ball4 is carrying right.  Left is a gripper.  Left is free.  Right is a gripper.  Robby is at roomb.  Room rooma Room roomb<|im_end|>
<|im_start|>assistant
<action>move from roomb to rooma</action><|im_end|>
<|im_start|>user
Ball1 is a ball.  Ball1 is at roomb.  Ball2 is a ball.  Ball2 is at roomb.  Ball3 is a ball.  Ball3 is at rooma.  Ball4 is a ball.  Ball4 is carrying right.  Left is a gripper.  Left is free.  Right is a gripper.  Robby is at rooma.  Room rooma Room roomb<|im_end|>
<|im_start|>assistant
<action>drop ball4 rooma right</action><|im_end|>
\end{lstlisting}
}
\end{tcolorbox}

\begin{tcolorbox}[notitle, sharp corners, breakable, colframe=Periwinkle, colback=white, 
       boxrule=3pt, boxsep=0.5pt, enhanced, 
       shadow={5pt}{-5pt}{0pt}{opacity=1,mygrey},
       title={Example trajectory on TriviaQA},]
       {\scriptsize
\begin{lstlisting}[
    language={},
    basicstyle=\ttfamily\scriptsize,
    breaklines=true,
    columns=fullflexible,
    numbers=none,
    backgroundcolor=\color{white},
    escapeinside={(*@}{@*)},
]
<|im_start|>user
In 2006, who was the first football player in World Cup history to win the "Best Young Player Award"?<|im_end|>
<|im_start|>assistant
<search> first football player to win Best Young Player Award at World Cup 2006 </search><|im_end|>
<|im_start|>user
Doc 1(Title: "FIFA World Cup awards") specific award for their position, they are still eligible for the Golden Ball as well, as when Oliver Kahn was awarded in 2002. In the event of a tie, the Golden Glove Award goes to the goalkeeper who progressed furthest in the competition. The next tiebreakers are saves made, then minutes played. Although the Golden Glove Award was first awarded in 1994, every All-Star Team in World Cups prior to 1998 except 1990 included only one goalkeeper. The ""Best Young Player"" award was awarded for the first time at the 2006 World Cup in Germany and given to Germany's Lukas
Doc 2(Title: "2006 Ballon d'O"") Netherlands, New Zealand, Norway, Poland, Portugal, Republic of Ireland, Romania, Russia, San Marino, Scotland, Serbia, Slovakia, Slovenia, Spain, Sweden, Switzerland, Turkey, Ukraine and Wales. Each picked a first (5pts), second (4pts), third (3pts), fourth (2pts) and fifth choice (1pt). Cannavaro was the fifth Italian player to win the award after Omar Sivori (1961), Gianni Rivera (1969), Paolo Rossi (1982) and Roberto Baggio (1993). Gianluigi Buffon (Italy) was the top ranked goalkeeper in the list, in second place; Ronaldinho (Brazil) was the top ranked midfielder, in fourth place, while Thierry Henry (France) was the top-ranked forward, in third place. The following
Doc 3(Title: "2006 FIFA World Cup") goals were scored by 110 players, with four of them credited as own goals. FIFA's Technical Study Group (TSG) also granted a Man of the Match award to one player in each match. Italy's Andrea Pirlo won the most Man of the Match awards, with three in total. Miroslav Klose, Agustin Delgado, Arjen Robben, Ze Roberto, Alexander Frei, Michael Ballack, and Patrick Vieira each received two awards. The All-star team is a squad consisting of the 23 most impressive players at the 2006 World Cup, as selected by FIFA's Technical Study Group. The team was chosen from a shortlist of<|im_end|>
<|im_start|>assistant
<search>first football player to win Best Young Player Award at 2006 World Cup</search><|im_end|>
<|im_start|>user
Doc 1(Title: "FIFA World Cup awards") specific award for their position, they are still eligible for the Golden Ball as well, as when Oliver Kahn was awarded in 2002. In the event of a tie, the Golden Glove Award goes to the goalkeeper who progressed furthest in the competition. The next tiebreakers are saves made, then minutes played. Although the Golden Glove Award was first awarded in 1994, every All-Star Team in World Cups prior to 1998 except 1990 included only one goalkeeper. The ""Best Young Player"" award was awarded for the first time at the 2006 World Cup in Germany and given to Germany's Lukas
Doc 2(Title: "2006 Ballon d'Or") Netherlands, New Zealand, Norway, Poland, Portugal, Republic of Ireland, Romania, Russia, San Marino, Scotland, Serbia, Slovakia, Slovenia, Spain, Sweden, Switzerland, Turkey, Ukraine and Wales. Each picked a first (5pts), second (4pts), third (3pts), fourth (2pts) and fifth choice (1pt). Cannavaro was the fifth Italian player to win the award after Omar Sivori (1961), Gianni Rivera (1969), Paolo Rossi (1982) and Roberto Baggio (1993). Gianluigi Buffon (Italy) was the top ranked goalkeeper in the list, in second place; Ronaldinho (Brazil) was the top ranked midfielder, in fourth place, while Thierry Henry (France) was the top-ranked forward, in third place. The following
Doc 3(Title: "2006 FIFA World Cup") goals were scored by 110 players, with four of them credited as own goals. FIFA's Technical Study Group (TSG) also granted a Man of the Match award to one player in each match. Italy's Andrea Pirlo won the most Man of the Match awards, with three in total. Miroslav Klose, Agustin Delgado, Arjen Robben, Ze Roberto, Alexander Frei, Michael Ballack, and Patrick Vieira each received two awards. The All-star team is a squad consisting of the 23 most impressive players at the 2006 World Cup, as selected by FIFA's Technical Study Group. The team was chosen from a shortlist of<|im_end|>
<|im_start|>assistant
<answer>Lukas Podolski</answer><|im_end|>
\end{lstlisting}
}
\end{tcolorbox}

\end{document}